\documentclass[fleqn,10pt]{wlscirep}
\usepackage[utf8]{inputenc}
\usepackage[T1]{fontenc}
\usepackage{amssymb}
\usepackage{booktabs}

\title{Simulating Three-dimensional Turbulence with Physics-informed Neural Networks}

\author[1,*]{Sifan Wang}
\author[2]{Shyam Sankaran}
\author[3]{Xiantao Fan}
\author[4]{Panos Stinis}
\author[2,*]{Paris Perdikaris}
\affil[1]{Institution for Foundation of Data Science, Yale University, New Haven, CT 06520}
\affil[2]{Department of Mechanical Engineering
  and Applied Mechanics,
  University of Pennsylvania,
  Philadelphia, PA 19104}
  \affil[3]{Sibley School of Mechanical and Aerospace Engineering, Cornell University, Ithaca, NY, 14850}
\affil[4]{Advanced Computing, Mathematics and Data Division, Pacific Northwest National Laboratory,
Richland, WA, 99354}
\affil[*]{sifan.wang@yale.edu, pgp@seas.upenn.edu}

\begin{abstract}
Turbulent fluid flows are among the most computationally     demanding problems in science, requiring enormous computational resources that become prohibitive at high flow speeds. Physics-informed neural networks (PINNs) represent a radically different approach that trains neural networks directly from physical equations rather than data, offering the potential for continuous, mesh-free solutions. Here we show that appropriately designed PINNs can successfully simulate fully turbulent flows in both two and three dimensions, directly learning solutions to the fundamental fluid equations without traditional computational grids or training data. Our approach combines several algorithmic innovations including adaptive network architectures, causal training, and advanced optimization methods to overcome the inherent challenges of learning chaotic dynamics. Through rigorous validation on challenging turbulent problems, we demonstrate that PINNs accurately reproduce key flow statistics including energy spectra, kinetic energy, enstrophy, and Reynolds stresses. Our results demonstrate that neural equation solvers can handle complex chaotic systems, opening new possibilities for continuous turbulence modeling that transcends traditional computational limitations.
\end{abstract}
\begin{document}

\flushbottom
\maketitle
%
%
\thispagestyle{empty}

\section*{Introduction}
Turbulence represents one of the last great unsolved problems in classical physics. Despite governing everything from the coffee swirling in your cup to the formation of galaxies, turbulent flows remain notoriously difficult to predict. This difficulty stems from their fundamentally chaotic nature: tiny changes in initial conditions can lead to dramatically different outcomes, while energy cascades across multiple scales create a computational challenge spanning orders of magnitude in both space and time.

The computational challenge is staggering. Conventional approaches such as direct numerical simulation (DNS) \cite{kim1987turbulence,moin1998direct} and large-eddy simulation (LES) \cite{lesieur1996new,piomelli1999large} have advanced our understanding considerably, but they remain prohibitively expensive at the high flow speeds encountered in most real-world applications. DNS requires resolving all relevant scales of motion on extremely fine computational grids, while LES introduces modeling assumptions to reduce cost but sacrifices accuracy. Both approaches scale poorly with increasing flow complexity, creating a fundamental barrier to progress in many areas of science and engineering.

This computational bottleneck has motivated the search for radically different approaches that can transcend the limitations of traditional grid-based methods. Physics-informed neural networks (PINNs) \cite{raissi2019physics, lagaris1998artificial} represent one such paradigm shift. Unlike conventional machine learning approaches that learn from data, PINNs embed the fundamental governing equations directly into the neural network training process, enabling them to learn continuous solutions to partial differential equations without requiring vast datasets or explicit spatial discretization.

Since their introduction, PINNs have shown promise  across diverse applications in fluid mechanics \cite{raissi2020hidden,almajid2022prediction,eivazi2022physics,cao2024surrogate}, heat transfer \cite{xu2023physics,bararnia2022application,gokhale2022physics}, bio-engineering \cite{kissas2020machine,zhang2023physics,caforio2024physics}, and materials science \cite{zhang2022analyses,jeong2023physics,hu2024physics}. However, their application to turbulent flows has remained elusive\cite{sarker2025quantasio,khademi2025simulation,khademi2025physics,jin2021nsfnets,botarelli2025using}. The combination of strong nonlinearity, extreme sensitivity to initial conditions, and multiscale dynamics presents a perfect storm of challenges that have historically caused PINN training to fail in turbulent regimes. The chaotic nature of turbulence, with its sensitive dependence on initial conditions and broad spectrum of active scales, pushes neural network optimization to its limits.

Here, we demonstrate that these fundamental challenges can be overcome. Through a combination of architectural innovations, advanced training strategies, and careful optimization techniques, we show for the first time that PINNs can successfully simulate fully developed turbulent flows in both two and three dimensions at high Reynolds numbers. Our framework achieves stable and accurate solutions across three canonical turbulence benchmarks: two-dimensional Kolmogorov flow, the three-dimensional Taylor–Green vortex, and turbulent channel flow. The resulting solutions reproduce key turbulence statistics -- including energy spectra, kinetic energy and enstrophy evolution, and Reynolds stress profiles -- at levels comparable to high-fidelity spectral solvers.

Our approach integrates several key innovations \cite{wang2023expert}: a physics-informed residual adaptive network (PirateNet) architecture that enables deep networks while mitigating spectral bias \cite{wang2024piratenets}; causal training strategies that respect temporal causality \cite{wang2022respecting}; self-adaptive loss weighting that balances competing optimization objectives \cite{wang2022and, wang2021understanding}; quasi-second-order optimization methods that navigate the complex loss landscapes inherent to turbulent systems \cite{wang2025gradient}; and time-windowing with transfer learning  \cite{liu2023adaptive,penwarden2023unified}. Together, these components enable robust training of large-scale PINNs for high-Reynolds-number turbulent flows.

While our method does not yet surpass state-of-the-art spectral solvers in computational efficiency, it establishes an important proof of concept: turbulence can be learned directly from governing equations using neural networks, offering continuous, mesh-free solutions that are not constrained by traditional discretization limitations. This work opens new possibilities for hybrid simulation approaches, data assimilation strategies, and scientific discovery in fluid dynamics and beyond.

\begin{figure}[htbp]
    \centering
    \includegraphics[width=1.0\linewidth]{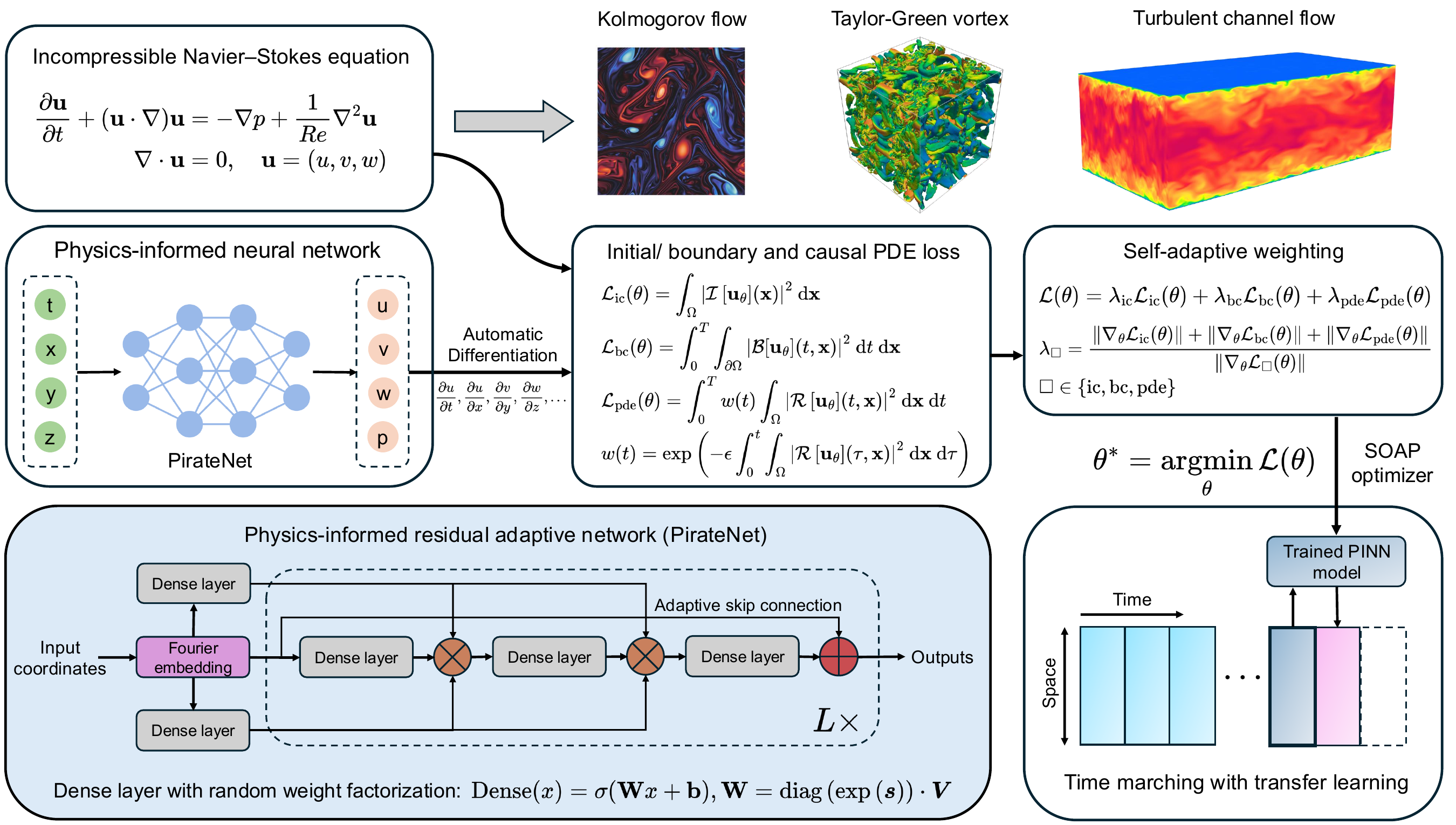}    \caption{{\em Physics-informed neural network framework for solving the incompressible Navier-Stokes equations.}  The PINN model takes spatiotemporal coordinates as input and predicts the velocity and pressure fields as $\mathbf{u}_\theta = (u_\theta, v_\theta, w_\theta, p_\theta)$, with $\theta$ denoting the network parameters. The network is trained by minimizing a composite loss that penalizes violations of PDE constraints and initial/boundary conditions, with PDE residuals computed via automatic differentiation. The framework integrates several key innovations: (a) a physics-informed residual adaptive network (PirateNet) that mitigates spectral bias and supports deep architectures; (b) a causal loss formulation that enforces temporal causality during training; (c) a self-adaptive weighting strategy that dynamically balances contributions from different loss terms; (d) a quasi-second-order optimizer (SOAP) that addresses conflicting gradient directions; and (e) a time-marching scheme with transfer learning, where the trainable parameters of models for later time windows are initialized using the converged model parameters for earlier time windows. These components collectively enable robust and efficient training of large-scale PINNs, making them viable for simulating high Reynolds number turbulent flows.}
    \label{fig:master_figure}
\end{figure}

\section*{Results}

We aim to solve the nondimensional incompressible Navier–Stokes equations in the velocity–pressure formulation using PINNs. As illustrated in Figure \ref{fig:master_figure}, the PINN approximates the velocity and pressure fields as neural network outputs, and computes the residuals of the governing equations via automatic differentiation with respect to the input coordinates. The network is trained by minimizing a composite loss that penalizes the PDE residuals, as well as deviations from the initial and boundary conditions.

We employ a suite of architectural and algorithmic innovations : (i) the PirateNet architecture \cite{wang2024piratenets}, which enables the use of large-capacity networks capable of capturing fine-scale turbulent structures; (ii) self-adaptive loss balancing \cite{wang2021understanding, wang2023expert}, which stabilizes training by dynamically adjusting the contributions of different loss terms; (iii) causal training \cite{wang2022respecting} and time-marching strategies \cite{wight2020solving,krishnapriyan2021characterizing} that facilitate convergence over long time horizons; (iv) the SOAP optimizer \cite{vyas2024soap,wang2025gradient} that is particularly well-suited for ill-conditioned PDE loss landscapes and mitigating gradient conflicts at scale; and (v) transfer learning which enables better initialization for the trainable parameters in each time window\cite{liu2023adaptive,penwarden2023unified}. Together, these components enable stable and robust training of PINNs for fully turbulent flows (indicated by their high-Reynolds-number, which is a nondimensional quantity characterizing the complexity of the flow). 

We evaluate our framework on three canonical benchmarks that test distinct aspects of turbulent behavior: the two-dimensional Kolmogorov flow to assess energy and enstrophy cascades in periodic domains, the three-dimensional Taylor–Green vortex to examine kinetic energy and enstrophy evolution in developing turbulence, and the three-dimensional turbulent channel flow to analyze velocity profiles and Reynolds stresses in near-wall regions. A  description of the methodology is provided in the Methods section, details about the training procedure and full hyperparameter settings are included in Appendix \ref{appendix:training} and additional visualizations and diagnostics can be found in Appendix \ref{appendix:visualizations}.

\subsection*{2D Turbulent Kolmogorov Flow}

\begin{figure}[htbp]
    \centering
    \includegraphics[width=1.0\linewidth]{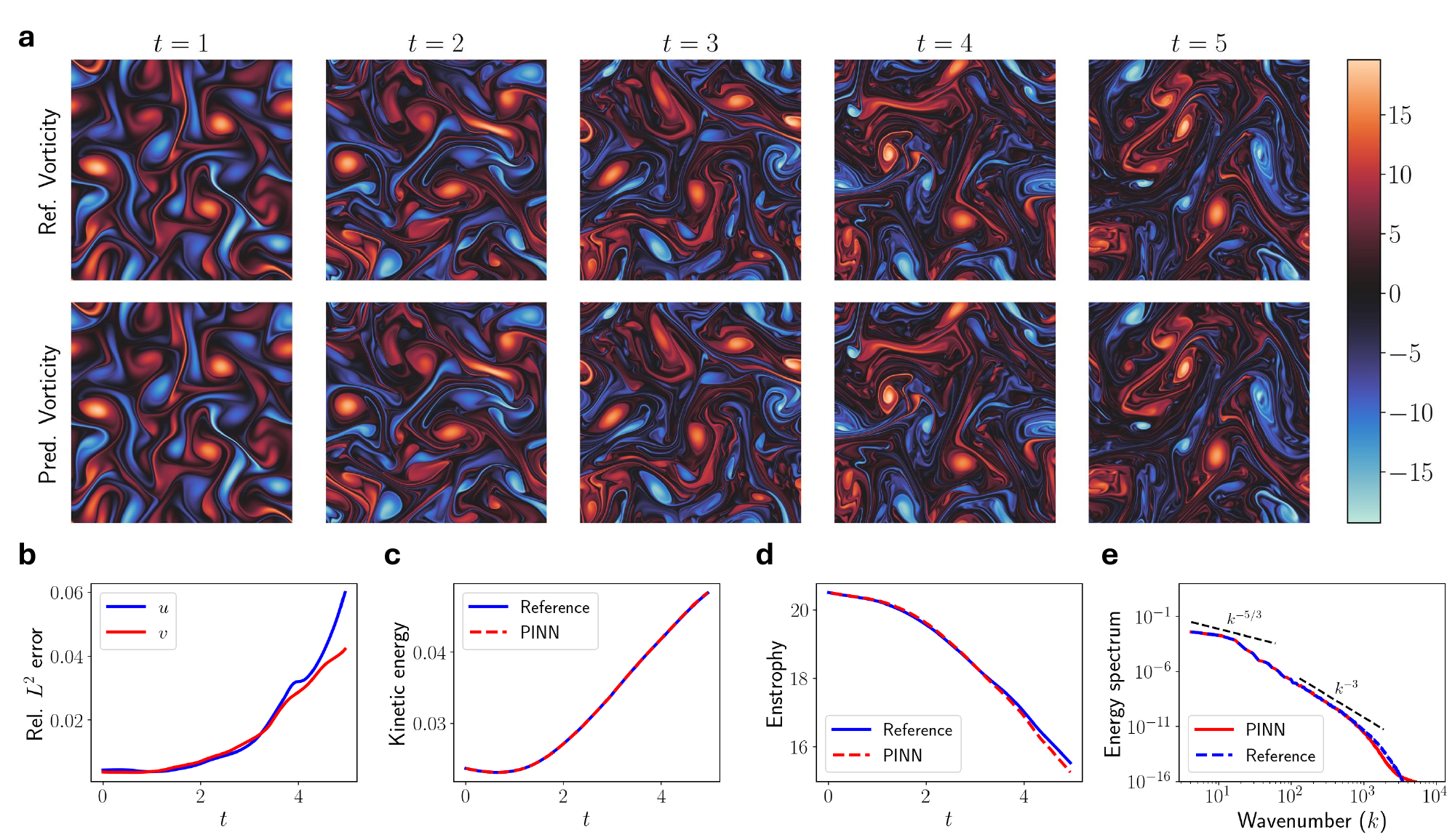}
    \caption{{\em 2D Turbulent Kolmogorov Flow ($\text{Re}=10^6$).} (a) Vorticity snapshots at selected time instances comparing predictions with reference DNS, computed using a pseudo-spectral method on a $2048 \times 2048$ uniform grid. The predicted fields closely match the DNS, accurately capturing fine-scale vortical structures and their temporal evolution. (b) Temporal evolution of the relative $L^2$ velocity error (spatially averaged), demonstrating sustained accuracy over the time domain.  (c–d) Comparison of kinetic energy and enstrophy evolution between PINN predictions and DNS results, showing good agreement in the preservation of key physical quantities. (e) Comparison of the energy spectrum at the final time between PINN predictions and DNS, illustrating accurate recovery of both the inverse energy cascade (\(k^{-5/3}\)) and the forward enstrophy cascade (\(k^{-3}\)).
    }
    \label{fig:kf_combined}
\end{figure}

We begin by evaluating our framework on the two-dimensional Kolmogorov flow, a periodic shear-driven system that serves as a standard benchmark for studying turbulence \cite{platt1991investigation,chandler2013invariant}. Despite its relatively simple setup, the flow exhibits nonlinear instabilities, an inverse energy cascade and a forward enstrophy cascade, and the formation of long-lived coherent structures\cite{doering1995applied} -- posing a meaningful challenge for data-free learning methods. We train the PINN on this system to assess its ability to reproduce key features of the flow, including vorticity evolution, energy spectra, and long-time statistical behavior.

The system is governed by the two-dimensional incompressible Navier-Stokes equations on the periodic domain $\Omega=[0,1]^2$, driven by a steady external forcing $\mathbf{f}(x, y)=\left[0.1 \sin (4 \pi y), 0 \right]$ 
which injects energy at wavenumber $k=2$. We  initialize the flow with a randomly generated velocity field and evolve it to a final time of 
 of $T=5$ at a Reynold number of $Re=10^6$.
At this regime, the system develops rich spatiotemporal dynamics characterized by an inverse energy cascade toward large-scale structures and a forward enstrophy cascade toward small scales. These dual cascading processes create a wide separation of active scales and complex long-range interactions, providing an informative test of the accuracy and stability of physics-informed neural networks.

We benchmark the PINN predictions against high-resolution DNS results obtained using a pseudo-spectral method on a $2048 \times 2048$ grid. Figure~\ref{fig:kf_combined} summarizes the comparison. Panel (a) shows that the predicted vorticity fields closely agree with the DNS across time, capturing both fine-scale features and large-scale structures. This agreement is further quantified in panel (b), where the relative $L^2$ error of the velocity field remains consistently low, indicating stable and accurate long-term predictions.
Panels (c) and (d) display the evolution of kinetic energy and enstrophy, respectively, both of which are in good agreement with the DNS, confirming the model’s ability to conserve key turbulence statistics. Finally, panel (e) presents the energy spectrum at $T = 5$, where the PINN accurately recovers both the $k^{-5/3}$ scaling of the inertial range and the $k^{-3}$ slope of the enstrophy cascade. This result highlights the model’s capacity to resolve nonlinear multiscale interactions across a wide range of wavenumbers.

\subsection*{3D Taylor-Green Vortex}

Having demonstrated our framework's capability in two dimensions, we now extend the evaluation to three-dimensional flows where vortex stretching and energy cascade mechanisms become fundamentally more complex. The Taylor–Green vortex (TGV) provides an ideal test case, featuring a well-defined transition from laminar initial conditions to fully developed turbulence through nonlinear vortex interactions \cite{brachet1983small,drikakis2007simulation}.

We consider the incompressible Navier-Stokes equations initialized with the  TGV configuration at Reynolds number $\operatorname{Re}=1,600$. The initial velocity field is defined as
\begin{align}
    u(x, y, z, 0) & =\sin (x) \cos (y) \cos (z), \\
v(x, y, z, 0) & =-\cos (x) \sin (y) \cos (z), \\
w(x, y, z, 0) & =0,
\end{align}
on a triply periodic domain $[0,2 \pi]^3$. The flow evolves through vortex stretching, roll-up, and eventual breakdown, leading to a fully developed turbulent state. Our objective is to simulate the flow evolution up to final time $T=10$.

\begin{figure}[htbp]
    \centering
    \includegraphics[width=1.0\linewidth]{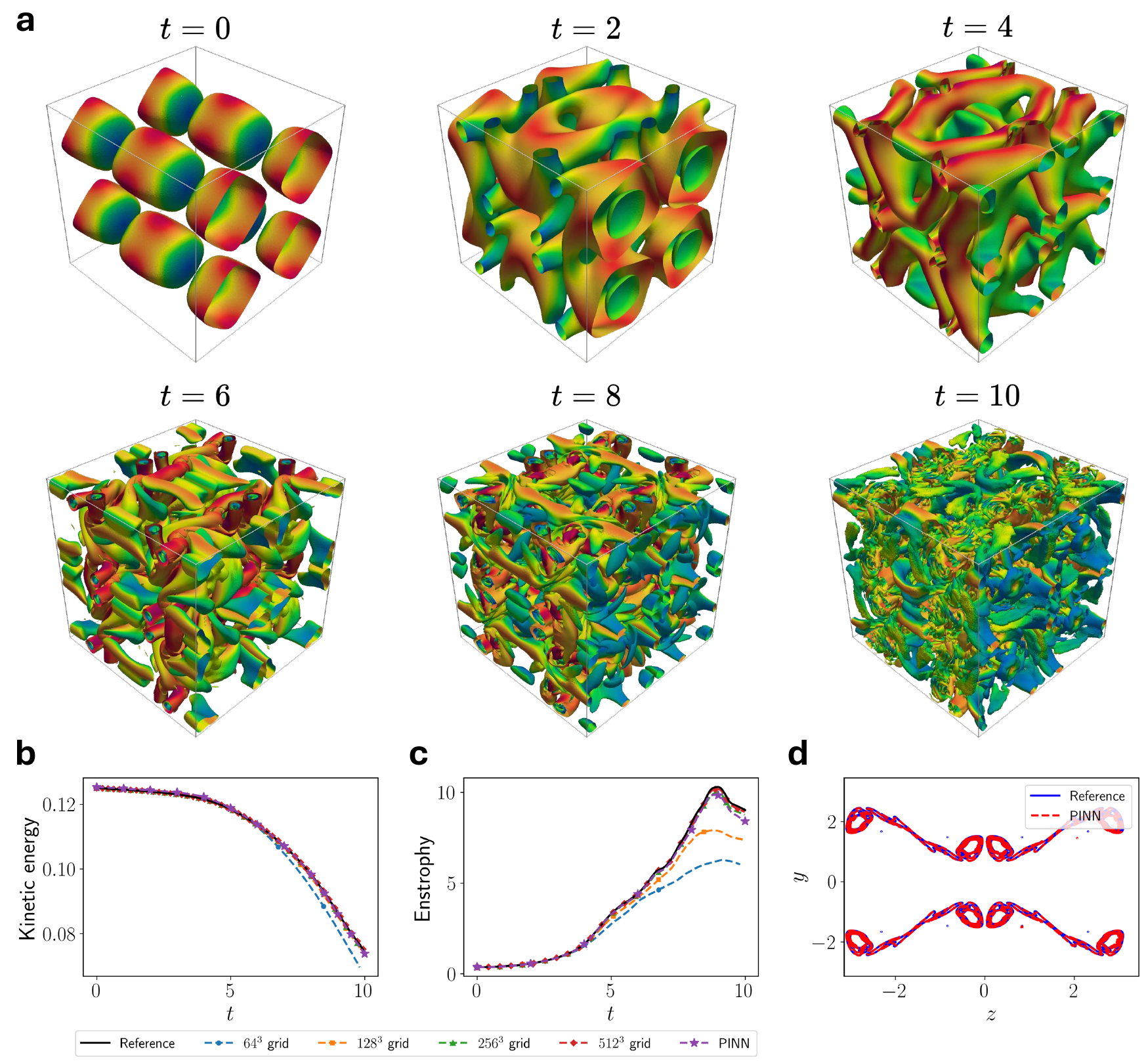}
    \caption{{\em Taylor-Green Vortex (Re=1600).}
    (a) Evolution of the iso-surfaces of the Q-criterion $(Q=0.1)$ at different time snapshots, predicted by PINNs and colored by the non-dimensional velocity magnitude.    
    (b–c) Temporal evolution of spatially averaged kinetic energy and enstrophy, comparing PINN predictions against a pseudo-spectral DNS (resolution \(512^3\)) and 8th-order finite difference solvers at various resolutions (\(64^3\)–\(512^3\)). The PINN achieves accuracy comparable to high-order solvers at moderate resolution and captures key dynamical features of the flow. (d) Comparison of the iso-contours of the dimensionless vorticity norm on the periodic face $x=-\pi$ at $t=8$.}
    \label{fig:tgv_combined}
\end{figure}

We validated our predictions against high-fidelity DNS data \cite{hiocfd2013}, which was computed using a dealiased pseudo-spectral method
on a $512^3$ grid with third-order Runge-Kutta time integration and timestep $\Delta t = 1.0 \times 10^{-3}$. 
Figure\ref{fig:tgv_combined}a presents the predicted isosurfaces of the Q criterion colored by the velocity magnitude.
These visualizations reveal that our PINN  captures the intricate vortex roll-up and breakdown dynamics characteristic of the laminar-turbulent transition. Quantitative validation through the evolution of spatially averaged kinetic energy and enstrophy (Figure\ref{fig:tgv_combined}b and Figure\ref{fig:tgv_combined}c, respectively) demonstrates strong statistical agreement with the DNS throughout the simulation period. Specifically, our model successfully reproduces both the overall energy decay profile and the characteristic enstrophy peak, which is associated with intense vortex stretching and the generation of small-scale structures. Furthermore, spatial validation via vorticity magnitude contours at t=8 (Figure\ref{fig:tgv_combined}d) further confirms the framework's ability to resolve fine-scale turbulent structures with high fidelity.

While our method does not yet surpass the accuracy of state-of-the-art spectral solvers, it performs comparably to an 8th-order finite difference method at a resolution of $256^3$. This demonstrated capability to accurately simulate complex three-dimensional turbulent transitions, capturing both global energy dynamics and local vortical structures, indicates the strong potential of neural PDE solvers  for practical applications.

\subsection*{3D Turbulent Channel Flow}
Building on our successful simulation of the TGV problem, we proceed to assess the framework's performance in three-dimensional turbulent channel flow. This prototypical wall-bounded system represents a significantly more stringent test for numerical methods. Unlike the periodic or isotropic flows previously considered, channel flow is characterized by sharp gradients near solid boundaries, inherent anisotropy, and nontrivial momentum transport. For a rigorous evaluation of predicted turbulence statistics, we specifically focus on the widely adopted benchmark case at a friction Reynolds number of $\mathrm{Re}_\tau=395$, for which extensive DNS statistics data are available \cite{del2004scaling,hoyas2008reynolds}.

We simulate incompressible channel flow within a computational domain adopting the standard configuration: $L_x=2 \pi$ in the streamwise direction, $L_z=\pi$ in the spanwise direction, and a wall-normal height of $L_y=2$. No-slip boundary conditions are applied at the channel walls ($y= \pm 1$), with periodicity enforced in the streamwise ($x$) and spanwise ($z$) directions. We set $u_\tau = 0.09875$ and the
 flow is driven by a constant pressure gradient of $d p / d x= u_\tau^2$, with the kinematic viscosity set to $\nu=0.00025$. Importantly, our PINN simulation  uses a snapshot of a turbulent DNS field as the initial condition. Our goal is to examine if PINNs can maintain and accurately reproduce the established turbulence statistics for this system over the time interval $t\in [0,10]$.

\begin{figure}[htbp]
    \centering
    \includegraphics[width=1.0\linewidth]{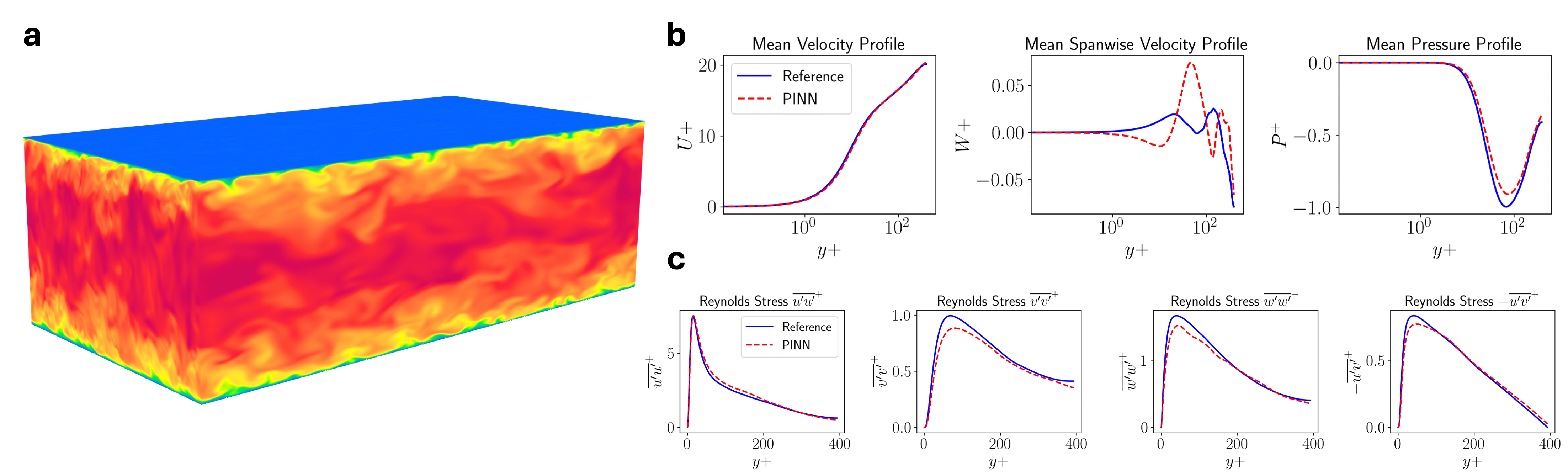}
    \caption{{\em Turbulent channel flow ($\text{Re}_\tau \sim 395$).}
    (a) Instantaneous snapshot of the predicted velocity norm, illustrating characteristic near-wall streaks and large-scale turbulent structures.
    (b) Comparison of the mean streamwise and spanwise velocity profile and pressrure profile
    between PINN predictions and DNS data, normalized in wall units.
    (c)  Comparison of Reynolds stress components ($\overline{u^{\prime}u^{\prime}}^{+}$, $\overline{v^{\prime}v^{\prime}}^{+}$, $\overline{w^{\prime}w^{\prime}}^{+}$, and $\overline{u^{\prime} v^{\prime}}^{+}$) between PINN predictions and DNS data, normalized by the respective friction velocity $u_\tau$. 
    }
    \label{fig:tcf_combined}
\end{figure}

Figure \ref{fig:tcf_combined} shows the framework's performance on the turbulent channel flow benchmark, compared against the reference DNS data\cite{del2004scaling,hoyas2008reynolds, moser1999direct}.
Figure \ref{fig:tcf_combined}a displays a representative instantaneous snapshot of the predicted velocity norm, which clearly captures the characteristic near-wall streaks and large-scale turbulent structures inherent to wall-bounded turbulence. Beyond this qualitative assessment, we quantify the learned statistics against reference DNS data from \cite{del2004scaling,hoyas2008reynolds, moser1999direct}.

Figure \ref{fig:tcf_combined}b shows the mean streamwise and spanwise velocity profiles together with the mean pressure distribution, all normalized in wall units. The PINN predictions closely follow the DNS data across the viscous sublayer, buffer layer, and logarithmic region, demonstrating the model’s ability to recover canonical mean flow behavior. Some discrepancies arises in the spanwise velocity component: because its magnitude is relatively small, the PINN struggles to capture it with the same accuracy as the streamwise counterpart.
Figure \ref{fig:tcf_combined}c presents the Reynolds stress components ${\overline{u^{\prime} u^{\prime}}}^{+}$, ${\overline{v^{\prime} v^{\prime}}}^{+}$, ${\overline{w^{\prime} w^{\prime}}}^{+}$, and ${\overline{u^{\prime} v^{\prime}}}^{+}$, normalized by the friction velocity $u_\tau$. Despite minor deviations, the predicted profiles show good agreement with DNS data in both magnitude and wall-normal distribution, successfully capturing the anisotropy of turbulent fluctuations. 

Further validation is provided in Figures \ref{fig:tcf_mean_vorticity}–\ref{fig:energy_spectra_ww}, which present the mean profiles of the RMS vorticity components and the energy spectra of the velocity components at different wall-normal locations $y^+$. The vorticity statistics confirm that the PINN accurately captures the near-wall peak and the wall-normal decay of all vorticity components, in close agreement with DNS data. The energy spectra further demonstrate that the predicted velocity fields reproduce the correct distribution of turbulent kinetic energy across scales, although the spectral amplitudes are slightly lower than the DNS reference. These results highlight that the PINN framework is capable of resolving both near-wall dynamics and multi-scale energy transfer, while still exhibiting a mild underprediction in spectral energy content.

\section*{Discussion}
This work demonstrates that physics-informed neural networks can successfully simulate fully turbulent flows across two and three dimensions at high Reynolds numbers. By combining architectural innovations with refined training strategies, we have shown that PINNs can accurately resolve canonical turbulence benchmarks including 2D Kolmogorov flow, the 3D Taylor–Green vortex, and 3D turbulent channel flow. These findings prove that continuous, mesh-free learning of chaotic, multiscale dynamics is achievable directly from the governing Navier–Stokes equations without relying on discretization grids or empirical turbulence models. Our success stems from addressing several fundamental challenges through the synergistic integration of scalable network architectures (PirateNet), adaptive loss weighting, causal training strategies, quasi-second-order optimization (SOAP), and transfer learning. These components work together to overcome the notorious difficulty of learning chaotic dynamics with neural networks.

Despite these advances, significant limitations remain. Our approach currently requires greater computational resources than conventional solvers (Appendix \ref{appendix:cost}), and while our solutions are physically consistent, they do not yet achieve the accuracy of state-of-the-art spectral methods. The computational overhead stems primarily from the iterative optimization process required to satisfy PDE constraints, which contrasts with the direct time-stepping of traditional methods. These limitations underscore the need for continued research in network architectures optimized for multiscale problems, more efficient training strategies, and rigorous benchmarking on progressively challenging problems. Importantly, turbulence provides an ideal testbed for the community to develop next-generation PINNs and other machine learning approaches for scientific computing. The field would benefit from moving beyond simple canonical problems towards more challenging benchmarks that expose the fundamental difficulties of learning complex physical systems.

Looking ahead, this work opens several promising research directions. Hybrid approaches that combine neural solvers with classical methods could leverage the strengths of both paradigms\cite{sanderse2024scientific,qadeer2025stabilizingpdemlcoupled}, while extensions to more complex scenarios -- including unsteady boundary conditions, anisotropic turbulence, compressible flows, and multiphysics systems represent natural next steps. Perhaps most significantly, the continuous, differentiable nature of PINN solutions makes them particularly well-suited for inverse problems, optimization, and control applications where traditional grid-based methods face fundamental limitations. Our results establish a foundation for physics-informed machine learning \cite{karniadakis2021physics} approaches that inherently respect physical laws while tackling the formidable complexity of real-world turbulent systems, marking a significant step toward more flexible, continuous computational fluid dynamics.

\section*{Methods}

The preceding results demonstrate that PINNs can  simulate high Reynolds-number turbulent flows. We now detail the computational methodological framework that enables these achievements.  

\subsection*{Problem Setup and Physics-informed Neural Networks}

We consider the problem of solving the incompressible Navier–Stokes equations, which govern the dynamics of viscous, incompressible fluid flows\cite{doering1995applied}. In non-dimensional form, the equations are 
\begin{align}
\frac{\partial \mathbf{u}}{\partial t}+\mathbf{u} \cdot \nabla \mathbf{u} & =-\nabla p+\frac{1}{\operatorname{Re}} \nabla^2 \mathbf{u}, \\
\nabla \cdot \mathbf{u} & =0,
\end{align}
where $\mathbf{u}(t, \mathbf{x})$ is the velocity field, $p(t, \mathbf{x})$ is the pressure, and Re is the Reynolds number. The equations are posed in a spatial domain $\Omega$ over a time interval $[0, T]$, and are supplemented with appropriate initial and boundary conditions defined at domain boundaries $\partial\Omega$. Due to the nonlinearity and the coupling between velocity and pressure fields, solving these equations -- especially in complex geometries or at high Reynolds numbers -- remains a significant challenge in computational fluid dynamics.

To solve the equations, we follow the conventional  formulation of PINNs \cite{raissi2019physics}
which integrates the governing physical laws directly into the neural network training process. The velocity and pressure fields are approximated by neural networks $\mathbf{u}_\theta(\mathbf{x}, t)$,  where  $\mathbf{u}_\theta = (u,v,w,p)$ and  $\theta$ represents the network parameters. The networks are trained by minimizing a composite loss function:
\begin{align}
    \mathcal{L}(\theta)=\underbrace{\int_{\Omega}\left|\mathcal{I}\left[\mathbf{u}_\theta\right](\mathbf{x})\right|^2 \mathrm{~d} \mathbf{x}}_{\mathcal{L}_{\mathrm{ic}}(\theta)}+\underbrace{\int_0^T \int_{\partial \Omega}\left|\mathcal{B}\left[\mathbf{u}_\theta\right](t, \mathbf{x})\right|^2 \mathrm{~d} \mathbf{x} \mathrm{~d} t}_{\mathcal{L}_{\mathrm{bc}}(\theta)}+\underbrace{\int_0^T  \int_{\Omega}\left|\mathcal{R}\left[\mathbf{u}_\theta\right](t, \mathbf{x})\right|^2 \mathrm{~d} \mathbf{x} \mathrm{~d} t}_{\mathcal{L}_{\mathrm{pde}}(\theta)},
\end{align}
where $\mathcal{I}[\cdot], \mathcal{B}[\cdot]$, and $\mathcal{R}[\cdot]$ denote the initial condition, boundary condition, and Navier-Stokes residual operators, respectively. Correspondingly, $\mathcal{L}_{\text {ic }}$ penalizes violations of the initial condition, $\mathcal{L}_{\text {bc }}$ enforces the boundary conditions, and $\mathcal{L}_{\text {pde }}$ enforces   the Navier-Stokes equations.

In practice, the network parameters are optimized using stochastic gradient descent and these continuous integrals are approximated using Monte Carlo sampling. Specifically, 
at each training iteration, collocation points are randomly sampled from the respective domains: the initial condition loss $\mathcal{L}_{\text {ic }}$ is computed from samples $\left\{\mathbf{x}_i^{\text {ic }}\right\}_{i=1}^{N_{\text {ic }}} \subset \Omega$, the boundary loss $\mathcal{L}_{\text {bc }}$ from points $\left\{\left(t_i^{\text {bc }}, \mathbf{x}_i^{\text {bc }}\right)\right\}_{i=1}^{N_{\text {bc }}} \subset[0, T] \times \partial \Omega$, and the PDE residual loss $\mathcal{L}_{\text {pde }}$ from interior collocation points $\left\{\left(t_i^{\text {pde }}, \mathbf{x}_i^{\text {pde }}\right)\right\}_{i=1}^{N_{\text {pde }}} \subset[0, T] \times \Omega$. The total loss is computed as the empirical average over these sample points,

A key advantage of PINNs is the use of automatic differentiation to compute PDE residuals directly from neural network outputs, enabling evaluation of spatial and temporal derivatives without explicit discretization schemes. This approach allows PINNs to operate on continuous domains without requiring mesh generation or traditional numerical solvers.

However, the applicability of conventional PINNs has been limited by critical challenges in training stability, scalability, and efficiency when applied to complex systems like turbulent flows. Specifically, standard PINNs often struggle to (i) accommodate the deep architectures necessary for resolving fine-scale structures \cite{wang2024piratenets}, (ii) navigate ill-conditioned optimization landscapes, (iii) mitigate training instabilities arising from competing loss terms \cite{wang2021understanding, wang2022and}; and (iv) respect the causal nature of time-dependent problems \cite{wang2022respecting}. To overcome these inherent limitations and enable stable, accurate learning in complex flow regimes, we introduce a suite of algorithmic and architectural innovations, detailed below.

\subsection*{Network Architecture}

Traditional PINNs exhibit degraded performance with larger and deeper neural network architectures, despite the necessity of such networks for approximating complex flows and sharp gradients. This counter-intuitive behavior stems from unsuitable initialization schemes in multi-layer perceptron architectures, leading to poor trainability of network derivatives and unstable minimization of the PDE residual loss \cite{wang2024piratenets}.

\paragraph{PirateNet.} To address these scalability challenges, we employ the PirateNet architecture \cite{wang2024piratenets} augmented with random weight factorization \cite{wang2022random}, which facilitates effective training of deep physics-informed neural networks while maintaining computational efficiency.

The architecture first transforms input coordinates $\mathbf{x}$ into a high-dimensional feature space using random Fourier features \cite{tancik2020fourier}:
\begin{align*}
    \Phi(\mathbf{x})= \begin{bmatrix}
    \cos (\mathbf{B x} ) \\
    \sin (\mathbf{B x} )
    \end{bmatrix},
\end{align*}
where $\mathbf{B} \in \mathbb{R}^{m \times d}$ has entries sampled i.i.d. from $\mathcal{N}(0, s^2)$ with user-specified $s > 0$.  We typically choose $d$ to be half the width of the neural network.
This embedding mitigates spectral bias in PINNs by improving the eigenfunction frequency of the Neural Tangent Kernel, thus enhancing the learning of high-frequency components and multiscale features \cite{wang2021eigenvector}.

The embedded coordinates are then processed through two dense layers that act as gates:
\begin{align*}
\mathbf{U} &= \sigma(\mathbf{W}_1 \Phi(\mathbf{x}) + \mathbf{b}_1  ), \quad
\mathbf{V} = \sigma(\mathbf{W}_2 \Phi(\mathbf{x}) + \mathbf{b}_2  ),
\end{align*}
where $\sigma$ is a point-wise activation function. 

Let $\mathbf{x}^{(1)} = \Phi(\mathbf{x})$ and $\mathbf{x}^{(l)}$ be the input to the $l$-th block ($1 \le l \le L$). Each block performs the following operations:
\begin{align}
    \mathbf{f}^{(l)} &= \sigma\big(\mathbf{W}^{(l)}_1 \mathbf{x}^{(l)} + \mathbf{b}^{(l)}_1\big)\,, &
    \mathbf{z}_1^{(l)} &= \mathbf{f}^{(l)} \odot \mathbf{U} + (1 - \mathbf{f}^{(l)}) \odot \mathbf{V}\,, \\
    \mathbf{g}^{(l)} &= \sigma\big(\mathbf{W}^{(l)}_2 \mathbf{z}_1^{(l)} + \mathbf{b}^{(l)}_2\big)\,, &
    \mathbf{z}_2^{(l)} &= \mathbf{g}^{(l)} \odot \mathbf{U} + (1 - \mathbf{g}^{(l)}) \odot \mathbf{V}\,, \\
    \mathbf{h}^{(l)} &= \sigma\big(\mathbf{W}^{(l)}_3 \mathbf{z}_2^{(l)} + \mathbf{b}^{(l)}_3\big)\,, &
    \mathbf{x}^{(l+1)} &= \alpha^{(l)} \mathbf{h}^{(l)} + (1 - \alpha^{(l)}) \mathbf{x}^{(l)}\,.
\end{align}

Each block consists of three dense layers with dual gating operations and an adaptive residual connection. The trainable $\alpha^{(l)}$ parameters control block nonlinearity: $\alpha^{(l)}=0$ yields an identity mapping, while $\alpha^{(l)}=1$ produces a fully nonlinear transformation. 
The final output of a PirateNet with $L$ residual blocks is given by
\begin{align}
    \mathbf{u}_{\mathbf{\theta}} = \mathbf{W}^{(L+1)} \mathbf{x}^{(L)}\,.
\end{align}
Importantly, we initialize $\alpha^{(l)}=0$ for all adaptive residual blocks, which initially makes the output a linear combination of first-layer embeddings. This initialization strategy mitigates training difficulties in deep PINNs by starting with an effectively shallow architecture and gradually increasing depth through learned $\alpha$ values during training.

\paragraph{Random weight factorization (RWF).}

We further enhance PirateNet's capabilities with Random Weight Factorization (RWF) \cite{wang2022random}, a drop-in replacement of dense layers that acts as learning rate booster to accelerate loss convergence. RWF decomposes each neuron's weight vector as:
\begin{align}
    \mathbf{w}^{(k, l)}= \exp(s^{(k, l)}) \cdot \mathbf{v}^{(k, l)},
\end{align}
where $k=1,\ldots,d_l$, $l=1,\ldots,L+1$, $\mathbf{w}^{(k, l)} \in \mathbb{R}^{d_{l-1}}$ is the $k$-th row of weight matrix $\mathbf{W}^{(l)}$, $s^{(k, l)} \in \mathbb{R}$ is a trainable scale factor, and $\mathbf{v}^{(k, l)} \in \mathbb{R}^{d_{i-1}}$. This factorization can be expressed in matrix form as:
\begin{align}
\mathbf{W}^{(l)}=\operatorname{diag}\left(\exp(\mathbf{s}^{(l)})\right) \cdot \mathbf{V}^{(l)}, \quad l=1,2, \ldots, L+1,
\end{align}
where $\mathbf{s}^{(l)} \in \mathbb{R}^{d_l}$ contains the log-scale parameters. The exponential reparameterization ensures that the scale factors are strictly positive.

Intuitively, RWF allows each neuron to effectively learn with its own self-adaptive learning rate, which helps to navigate the complex loss landscape more efficiently and overcome issues like spectral bias and poor initialization \cite{wang2022random}.

\paragraph{Exact enforcing of periodic boundary conditions.} 

We enforce the periodic boundary conditions as hard constraints, following the approach reported in Dong {\it et al.} \cite{dong2021method}.  For a function $u(\mathbf{x})$ which is periodic along an arbitrary axis, this is achieved by transforming the input coordinates using a Fourier feature embedding.

Specifically, for a domain periodic in a given direction with period $P$, we modify the input to the neural network for that coordinate $x_i$ to include $\cos \left(\omega_i x_i\right)$ and $\sin \left(\omega_i x_i\right)$, where $\omega_i=\frac{2 \pi}{P_i}$ is the frequency corresponding to the period $P_i$ in that direction. This generalizes directly to multiple periodic dimensions. For example, in a 3D domain with periodicity in $x, y, z$ directions, the spatial embedding becomes:
\begin{align}
    \mathbf{\gamma}(x, y, z)=\left[\cos \left(\omega_x x\right), \sin \left(\omega_x x\right), \cos \left(\omega_y y\right), \sin \left(\omega_y y\right), \cos \left(\omega_z z\right), \sin \left(\omega_z z\right)\right].
\end{align}
Any network $u_\theta(\mathbf{\gamma}(\mathbf{x}))$ using such an embedding will inherently satisfy the periodic boundary conditions for all derivatives. For time-dependent problems, time coordinates $t$ are simply concatenated with these spatial embeddings.

\subsection*{Training Protocol}
Training PINNs presents significant challenges, including conflicting gradients among loss components, ill-conditioned loss landscapes, and violations of temporal causality in time-dependent problems. These difficulties often lead to poor convergence, reduced solution accuracy, and physically inconsistent results. To address these fundamental issues, we employ several specialized techniques that enhance optimization stability and predictive accuracy.

\paragraph{Causal training.} Recent work \cite{wang2022respecting} illustrates that PINNs may violate temporal causality when solving time-dependent PDEs, often minimizing the PDE residuals at later times before accurately obtaining the solution for earlier times. To address this, we employ a causality-aware training approach\cite{wang2022respecting}, where the PDE residual loss is modified as:
\begin{align}
    & \mathcal{L}_{\text {pde }}(\theta)=\int_0^T w(t) \int_{\Omega}\left|\mathcal{R}\left[\mathbf{u}_\theta\right](t, \mathbf{x})\right|^2 \mathrm{~d} \mathbf{x} \mathrm{~d} t, \\
& w(t)=\exp \left(-\epsilon \int_0^t \int_{\Omega}\left|\mathcal{R}\left[\mathbf{u}_\theta\right](\tau, \mathbf{x})\right|^2 \mathrm{~d} \mathbf{x} \mathrm{~d} \tau\right).
\end{align}
The weight $w(t)$ decreases exponentially  with the cumulative residual loss from earlier times. This mechanism ensures that the network prioritizes learning the solutions  at earlier times before focusing on later ones, thereby enforcing temporal causality during the optimization process.

\paragraph{Self-adaptive weighting.}
A central challenge in training PINNs is the imbalance among loss components, which often reflects distinct physical scales and yields gradients of vastly different magnitudes. This imbalance can lead to unstable optimization and poor convergence.
To address this, we implement a self-adaptive learning rate annealing scheme \cite{wang2021understanding} that dynamically balances the weighted composite  loss:
\begin{align}
    \mathcal{L}(\mathbf{\theta}) =  \lambda_{ic} \mathcal{L}_{ic}(\mathbf{\theta}) + \lambda_{bc} \mathcal{L}_{bc}(\mathbf{\theta}) +  \lambda_r \mathcal{L}_r(\mathbf{\theta}), 
\end{align}
The global weights are computed as:
\begin{align}
    &\lambda_{\square}=\frac{\left\|\nabla_\theta \mathcal{L}_{\mathrm{ic}}(\theta)\right\|+\left\|\nabla_\theta \mathcal{L}_{\mathrm{bc}}(\theta)\right\|+\left\|\nabla_\theta \mathcal{L}_{\mathrm{pde}}(\theta)\right\|}{\left\|\nabla_\theta \mathcal{L}_{\square}(\theta)\right\|},
\end{align}
where $\square \in\{\mathrm{ic}, \mathrm{bc}, \mathrm{pde}\} $ and $\|\cdot\|$ denotes the $L^2$ norm.
Then we obtain
\begin{align}
  \| \lambda_{ic} \nabla_\theta \mathcal{L}_{ic} (\theta) \| =   \| \lambda_{bc} \nabla_\theta \mathcal{L}_{ic} (\theta) \| = \| \lambda_{r} \nabla_\theta \mathcal{L}_{ic} (\theta) \| = \| \nabla_\theta \mathcal{L}_{ic} (\theta) \| +  \| \nabla_\theta \mathcal{L}_{bc} (\theta) \|  +  \| \nabla_\theta \mathcal{L}_{r} (\theta) \|. 
\end{align}
This formulation equalizes the gradient norms of weighted losses, preventing bias toward any particular term during training. The weights are updated as running averages of their previous values, stabilizing stochastic gradient descent. Updates occur at user-specified intervals (typically every 100-1000 iterations), leading to minimal computational overhead.

\paragraph{SOAP optimizer.} As demonstrated by Wang et al. \cite{wang2025gradient}, PINNs frequently suffer from conflicting gradient directions, which can hinder optimization. To mitigate this issue and improve training efficiency, we employ SOAP, a recent optimizer introduced by Vyas et al. \cite{vyas2024soap}

SOAP operates by performing Adam optimization within an eigenspace constructed from the covariance matrix of gradients.
Specifically, at each optimization iteration $\tau$ for each layer's weight matrix $W_t$ and its corresponding gradient $G_t \in \mathbb{R}^{m \times n}$, SOAP maintains two covariance matrices using exponential moving averages:
\begin{align}
    & L_\tau=\beta_2 L_{\tau-1}+\left(1-\beta_2\right) G_\tau G_\tau^T ,\\
    & R_\tau=\beta_2 R_{\tau-1}+\left(1-\beta_2\right) G_\tau^T G_\tau.
\end{align}
These matrices are then eigendecomposed as $L_\tau=Q_L \Lambda_L Q_L^T$ and $R_\tau=Q_R \Lambda_R Q_R^T$. Here, $\Lambda_L$ and $\Lambda_R$ contain eigenvalues that capture the principal curvature directions of the loss landscape, providing crucial information for navigating the optimization space.

At each iteration $\tau$, SOAP updates each layer's weight matrix $W_\tau$ with a learning rate  $\eta$ using its corresponding gradient $G_\tau$ as follows:
\begin{enumerate}
    \item Project the weight and gradient onto the eigenspace: $\widetilde{W}_\tau = Q_L^T W_\tau Q_R, \widetilde{G}_\tau = Q_L^T G_\tau Q_R.$
    \item Apply the Adam update in the \emph{rotated} space:
    $\widetilde{W}_{\tau+1} = \widetilde{W}_{\tau} - \eta \, \operatorname{Adam}(\widetilde{G}_\tau).$
    \item Transform back to the original parameter space: $W_{\tau+1} = Q_L \widetilde{W}_{\tau+1} Q_R^T.$
\end{enumerate}
SOAP can be viewed as an approximation to the Gauss-Newton method \cite{wang2025gradient}. Consequently, SOAP actively mitigates conflicting gradient directions and preconditions the often ill-posed PINN loss landscape, leading to more stable and efficient training.

\paragraph{Time-marching and transfer learning}

To further enhance model performance and reduce optimization difficulty, we employ a time-marching strategy that decomposes the temporal domain into sequential windows. Rather than training a PINN model on the entire time domain simultaneously, we solve the problem progressively by  training a separate PINN model for each consecutive time interval $[t_i, t_{i+1}]$ for $i = 0, 1, \ldots, N-1$, where $t_0 = 0$ and $t_N = T$.

For each time window, we implement transfer learning by initializing the PINN parameters using the parameter values of the converged PINN model from the previous time interval \cite{liu2023adaptive,penwarden2023unified}. Specifically, the network parameters $\theta^{(i+1)}$ for time window $[t_{i+1}, t_{i+2}]$ are initialized with the optimized parameters $\theta^{(i)}_*$ from the preceding window $[t_i, t_{i+1}]$. The initial condition for each new time window is obtained from the PINN solution at the final time of the previous interval, ensuring continuity across temporal segments. Our experiments demonstrate that this time-marching strategy with transfer learning leads to significantly improved loss convergence and solution accuracy compared to training over the entire temporal domain simultaneously from scratch.


\section*{Acknowledgments}
The work of P.S., P.P. and S.S. was supported by the U.S. Department of Energy, Advanced Scientific Computing Research program, under the ``Resolution-invariant deep learning for accelerated propagation of epistemic and aleatory uncertainty in multi-scale energy storage systems, and beyond'' project (Project No. 81824). Pacific Northwest National Laboratory (PNNL) is a multi-program national laboratory operated for the U.S. Department of Energy (DOE) by Battelle Memorial Institute under Contract No. DE-AC05-76RL01830.

\section*{Author Contributions Statement}

S.W., P.S., and P.P. conceived the project and jointly designed the methodology. S.W. performed the experiments and analyzed the data. S.S. contributed to the implementation and assisted with experimental validation. All authors contributed to interpreting the results and to writing and revising the manuscript.

\section*{Code and Reproducibility}

All code used in this study, along with instructions for reproducing the experiments and analyses, will be made publicly available upon publication. Detailed documentation and scripts are provided to facilitate reproducibility of the results.

\section*{Competing interests}

The authors declare no competing interests.



\bibliography{reference}

\clearpage

\appendix

\section{Nomenclature}
\label{appendix:notations}

\begin{table}[htpb]
    \small
    \renewcommand{\arraystretch}{1.1}
\centering
\caption{Notation used throughout the paper.}
\label{tab:notation}
\begin{tabular}{c|l}
\toprule
\textbf{Symbol} & \textbf{Description} \\
\midrule
\multicolumn{2}{c}{\textbf{Physical Variables}} \\
$\mathbf{u}$, $p$ & Velocity and pressure fields \\
$t$, $\mathbf{x}$ & Time and spatial coordinates \\
$\Omega$, $T$ & Spatial domain and final time \\
$\text{Re}$, $\text{Re}_\tau$ & Reynolds and friction Reynolds numbers \\
$u_\tau$ & Friction velocity  \\
\midrule
\multicolumn{2}{c}{\textbf{Neural Networks}} \\
$\theta$ & Network parameters \\
$\mathbf{u}_\theta$ & Network output, denoting the physical fields $(u,v,w,p)$ \\
$\mathbf{W}^{(l)}$, $\mathbf{b}^{(l)}$ & Weight matrix and bias for layer $l$ \\
$\alpha^{(l)}$ & Trainable adaptive residual parameter \\
$L$ & Number of adaptive residual blokcs \\
$\Phi(\mathbf{x})$ & Random Fourier features \\
\midrule
\multicolumn{2}{c}{\textbf{Loss Functions}} \\
$\mathcal{L}$ & Total loss function \\
$\mathcal{L}_{\text{ic}}$, $\mathcal{L}_{\text{bc}}$, $\mathcal{L}_{\text{pde}}$ & Initial, boundary, and PDE losses \\
$\mathcal{I}[\cdot]$, $\mathcal{B}[\cdot]$, $\mathcal{R}[\cdot]$ & Initial, boundary, and residual operators \\
$\lambda_{\text{ic}}$, $\lambda_{\text{bc}}$, $\lambda_r$ & Adaptive loss weights \\
$w(t)$ & Causal training weight \\
$\epsilon$ & Causal tolerence \\
\midrule
\multicolumn{2}{c}{\textbf{Optimization}} \\
$G_\tau$ & Gradient matrix at iteration $\tau$ \\
$L_\tau$, $R_\tau$ & SOAP covariance matrices \\
$Q_L$, $Q_R$ & Eigenvector matrices of the covariance matrices \\
$\eta$ & Learning rate \\
$\mathbf{s}, \mathbf{V}$ & Random weight factorization \\
\midrule
\multicolumn{2}{c}{\textbf{Turbulence Diagnostics}} \\
$k$ & Wavenumber \\
$Q$ & Q-criterion \\
$U^+$ & Mean streamwise velocity (wall units) \\
$u'$, $v'$, $w'$ & Velocity fluctuations (wall units)\\
$u'_{rms}$, $v'_{rms}$, $w'_{rms}$ & Root-mean-square (rms) of velocity fluctuations     (wall units) \\
$\overline{u'u'}^+$, $\overline{v'v'}^+$, $\overline{w'w'}^+$, $\overline{u'v'}^+$  & Reynolds stress components (wall units)\\
$\omega_x$, $\omega_y$, $\omega_z$ & Vorticity fluctuations (wall units) \\
$\omega_x^{rms}$, $\omega_y^{rms}$, $\omega_z^{rms}$ & Root-mean-square (rms) of vorticity fluctuations(wall units) \\

\midrule
\multicolumn{2}{c}{\textbf{Sampling}} \\
$N_{\text{ic}}$, $N_{\text{bc}}$, $N_{\text{pde}}$ & Number of sample points for evaluating the initial/boundary and PDE losses\\
\bottomrule
\end{tabular}
\end{table}

\clearpage
\section{Metrics}
\label{appendix:metrics}

We detail the evaluation metrics used in our work to assess the accuracy and physical fidelity of the predicted flow fields.

\paragraph{Relative $L^2$ error.}
The relative $L^2$ error between the predicted solution $\tilde{\mathbf{u}}$ and the reference solution $\mathbf{u}$ is defined as
\begin{equation}
\text{Relative } L^2 \text{ Error} = \frac{\| \tilde{\mathbf{u}} - \mathbf{u} \|_2}{\| \mathbf{u} \|_2},
\end{equation}
which quantifies the overall discrepancy between the predicted and true velocity fields over the spatial domain.

\paragraph{Kinetic energy.}
The kinetic energy of the velocity field is computed as
\begin{equation}
\text{KE} = \frac{1}{2} \int_{\Omega} \|\mathbf{u}(\mathbf{x})\|^2 \, d\mathbf{x},
\end{equation}
providing a global measure of the flow's intensity. We report the relative error in kinetic energy compared to the reference data.

\paragraph{Kinetic energy spectrum.}
We compute the kinetic energy spectrum (energy spectrum of the velocity field) to analyze the distribution of energy across spatial scales. The Fourier transform $\hat{\boldsymbol{u}}(\mathbf{k})$ is used to calculate the spectrum:
\begin{equation}
E(k) = \sum_{|\mathbf{k}| = k} \frac{1}{2} |\hat{\boldsymbol{u}}(\mathbf{k})|^2,
\end{equation}
where $\mathbf{k}$ denotes the wavevector. This metric  captures 
the spectral decay of turbulence.

\paragraph{Enstrophy.}
Enstrophy measures the rotational content of the flow and is defined as
\begin{equation}
\text{Enstrophy} = \frac{1}{2} \int_{\Omega} \|\nabla \times \mathbf{u}(\mathbf{x})\|^2 \, d\mathbf{x}.
\end{equation}
It characterizes the presence of small-scale structures and vorticity in the flow.

\paragraph{Enstrophy spectrum.}
We compute the enstrophy spectrum (energy spectrum of the vorticity field) to analyze the distribution of rotational energy across spatial scales. The vorticity $\boldsymbol{\omega} = \nabla \times \mathbf{u}$ is first computed, and its Fourier transform $\hat{\boldsymbol{\omega}}(\mathbf{k})$ is used to calculate the spectrum:
\begin{equation}
E_{\omega}(k) = \sum_{|\mathbf{k}| = k} \frac{1}{2} |\hat{\boldsymbol{\omega}}(\mathbf{k})|^2,
\end{equation}
where $\mathbf{k}$ denotes the wavevector. This metric is particularly sensitive to small-scale structures and is useful for evaluating how well the model captures the fine-scale features and spectral decay of turbulence.

\paragraph{Mean velocity and pressure profiles (space--time average).}
For statistically steady turbulent channel flow, the mean velocity and pressure profiles are obtained by averaging over the streamwise (\(x\)), spanwise (\(z\)), and time (\(t\)) directions:
\begin{equation}
 \overline{u}_i (y)
= \frac{1}{T\,L_x L_z} \int_{t_0}^{t_0+T} \int_0^{L_x} \int_0^{L_z}
u_i(x,y,z,t)\, dz\, dx\, dt,
\qquad u_i \in \{u,v,w\},
\end{equation}
and
\begin{equation}
 \overline{p}(y)
= \frac{1}{T\,L_x L_z} \int_{t_0}^{t_0+T} \int_0^{L_x} \int_0^{L_z}
p(x,y,z,t)\, dz\, dx\, dt.
\end{equation}
These Reynolds (space--time) averages yield profiles that depend only on the wall-normal coordinate \(y\), providing a robust measure of how well the model captures the mean dynamics.

To facilitate comparison with reference DNS data, the mean quantities are expressed in wall units:
\[
U^+(y^+) = \frac{\overline{u}(y)}{u_\tau}, \qquad 
W^+(y^+) = \frac{\overline{w}(y)}{u_\tau}, \qquad 
P^+(y^+) = \frac{\overline{p}(y)}{\rho u_\tau^2},
\]
where
\[
y^+ = \frac{y u_\tau}{\nu}, 
\qquad 
u_\tau = \sqrt{\tau_w / \rho},
\]
with \(\tau_w\) the mean wall shear stress and \(\nu\) the kinematic viscosity.

\paragraph{Root mean square (RMS) velocity and vorticity fluctuations.}
The RMS velocity fluctuations for each component are defined as
\begin{equation}
u_{i}^{\text{rms}}(y) = \sqrt{ \overline{u_i'^2} }, 
\qquad 
u_i' = u_i - \overline{u_i},
\end{equation}
where the averaging is performed over the streamwise (\(x\)), spanwise (\(z\)), and time (\(t\)) directions:
\begin{equation}
\overline{u_i'^2}(y) 
= \frac{1}{T L_x L_z} \int_{t_0}^{t_0+T} \int_0^{L_x} \int_0^{L_z} 
\left( u_i(x,y,z,t) - \overline{u_i}(y) \right)^2 
\, dz \, dx \, dt.
\end{equation}
These profiles quantify the turbulence intensity and directional anisotropy of the flow.

Similarly, the RMS vorticity fluctuations are given by
\begin{equation}
\omega_{i}^{\text{rms}}(y) = \sqrt{ \overline{\omega_i^2} },
\end{equation}
where \(\omega_i \in \{\omega_x, \omega_y, \omega_z\}\) are the components of the vorticity field, and
\begin{equation}
\overline{\omega_i^2}(y) 
= \frac{1}{T L_x L_z} \int_{t_0}^{t_0+T} \int_0^{L_x} \int_0^{L_z} 
\omega_i^2(x,y,z,t) \, dz \, dx \, dt.
\end{equation}
These vorticity-based statistics highlight the strength of small-scale rotational structures and provide complementary insight into the near-wall dynamics.

\paragraph{Reynolds stress.}
The Reynolds stress components quantify turbulent momentum transport and are defined as
\begin{equation}
\overline{u'_i u'_j} = \overline{(u_i - \overline{u_i})(u_j - \overline{u_j})}.
\end{equation}
In turbulent flows such as channel flow, the $\overline{u'v'}$ component is especially important as it represents shear stress due to turbulence. Accurate modeling of Reynolds stresses is crucial for reproducing the correct momentum balance.
\paragraph{Velocity energy spectrum.} The energy spectrum for each velocity component at different wall-normal locations characterizes the distribution of kinetic energy across wavenumbers. For the velocity component $u_i$, the one-dimensional energy spectrum in the $x_i$-direction is computed as
\begin{equation}
E_{u_i u_i}(k_i, y) = \left\langle \sum_{k_j}  \hat{u}_i(k_i, k_j, y), \hat{u}_i^*(k_i, k_j, y) \right\rangle,
\end{equation}
where $k_i$ is the streamwise ($i=1$) or spanwise ($i=3$) wavenumber, $k_j$ is the complementary horizontal wavenumber, $\hat{u}_i$ denotes the Fourier coefficient of $u_i$, and $(\cdot)^*$ indicates the complex conjugate. The angle brackets $\left\langle \cdot \right\rangle$ denote ensemble averaging over multiple samples to obtain a statistically converged spectrum.

\section{Training protocols}
\label{appendix:training}

\begin{table}
\renewcommand{\arraystretch}{1.2}
\centering
\caption{{\em Unified hyperparameter configuration for all turbulence benchmarks.} RFF and RWF represent Random Fourier Features and Random Weight Factorization, respectively.}
\label{tab:hyper-parameters}
\begin{tabular}{l c}
\toprule
\textbf{Parameter} & \textbf{Value} \\
\midrule
\textbf{PirateNet Architecture} & \\
\midrule
Depth (residual blocks) & 2 \\
Width (hidden size) & 768 \\
Activation & Swish \\
RFF scale & $\mathcal{N}(0, 2)$ \\
RWF & $\mu{=}1.0, \sigma{=}0.1$ \\
\midrule
\textbf{Optimization} & \\
\midrule
Optimizer & SOAP \\
$\beta_1$, $\beta_2$ & 0.9, 0.999 \\
Eigenspace update frequency & 2 steps \\
Base learning rate & $10^{-3}$ \\
Warmup steps & 2,000 \\
Decay rate & 0.9 \\
Decay steps & 2,000 \\
\midrule
\textbf{Training} & \\
\midrule
Iterations (per time window) & $10^5$ \\
Batch size & 8,192 \\
Time window size & 0.1 \\ 
Loss weighting & Grad Norm (updated every 1,000 iters) \\
Causal tolerance & $\epsilon = 1.0$ \\
\bottomrule
\end{tabular}
\end{table}

We adopt a unified training protocols across all benchmark tasks, with problem-specific adaptations detailed in the respective sections below. 

All models are trained using a time-marching strategy with transfer learning across temporal windows. 
We use a time window size of 0.1 by default. The number of windows is chosen empirically based on problem complexity, with adjustments guided by preliminary observations of loss convergence. 
Within each window, we use PirateNet \cite{wang2024piratenets} as the network backbone, comprising two residual blocks (six layers total), a hidden size of 768, and \texttt{Swish} activations. Network weights are initialized via random weight factorization (RWF) \cite{wang2022random} using $\mu = 1.0$ and $\sigma = 0.1$. When applicable, periodic boundary conditions are enforced exactly following the method in \cite{dong2021method}, and Fourier features are sampled from a Gaussian distribution $\mathcal{N}(0, 2)$.

To ensure stable and effective training, we adopt a self-adaptive loss weighting scheme \cite{wang2021understanding, wang2023expert}, where weights are updated every 1,000 iterations using an exponential moving average. For causal training \cite{wang2022respecting, wang2023expert}, each time window is further divided into disjoint chunks, and causal weights are evaluated over these chunks with a tolerance of $\epsilon = 1.0$.

The model is trained by minimizing the composite loss over $10^5$ iterations using the SOAP optimizer, with parameters $\beta_1 = 0.9$, $\beta_2 = 0.999$, and eigenspace updates every $f = 2$ steps. The learning rate increases linearly to $10^{-3}$ in the first 22,000 steps, followed by an exponential decay with a factor of 0.9 every 2,000 steps. To further accelerate convergence, we employ the schedule-free method of \cite{defazio2024road} with momentum $\beta = 0.9$, together with gradient clipping (global norm of 1) to ensure training stability.

A complete summary of hyperparameters is provided in Table~\ref{tab:hyper-parameters}, based on configurations validated in prior work \cite{wang2023expert, wang2024piratenets}. These hyperparameters were also further tuned via extensive ablation studies, which we omit here for brevity.

\paragraph{Taylor-Green vortex.} 

For this benchmark, we use a time window size of 0.2 for $t \in [0, 8]$, and reduce it to 0.05 for later times as the flow becomes increasingly turbulent. Empirically, smaller time windows result in lower training loss and improved resolution of turbulence statistics such as enstrophy. These observations motivate future research into adaptive time window strategies, where the window size is adjusted dynamically based on the evolving complexity of the solution.

To further enhance solution accuracy, we adopt the multi-stage training scheme introduced by Wang et al. \cite{wang2024multi}, which addresses the inherent limitations of neural networks in achieving high
accuracy. In the context of PINNs, this method constructs the solution as a series expansion where each term is represented by a neural network. The neural network in the series expansion are trained sequentially to reduce the equation residuals left by preceding stages:
\begin{align}
u_c^{(n)}(t,\mathbf{x})= \sum_{j=0}^n \varepsilon_j u_j(t,\mathbf{x}),
\end{align}
where $u_j(x)$ denotes the neural network at stage $j$, and $\varepsilon_j$ is a prescribed scaling factor.

Training proceeds stage-by-stage, following Algorithm 4 from Wang et al. \cite{wang2024multi}:
\begin{enumerate}
\item Stage 0: Train the initial network $u_0(t,\mathbf{x})$ using standard PINN optimization to minimize the governing equation residuals and enforce initial/boundary conditions.

\item Stage $k+1$:
\begin{itemize}
    \item Train a new network \( u_{k+1}(t,\mathbf{x}) \) with a  magnitude prefactor \( \varepsilon_{k+1} \), by minimizing the physics-informed loss of the linearized PDEs. The non-linear PDEs are linearized around the current solution \( u_c^{(k)}(t,\mathbf{x}) \), yielding linear residual equations for \( u_{k+1}(t,\mathbf{x}) \), where the residual from stage \( k \) acts as the source term.
    \item Update the composite solution as:
    \begin{align}
        u_c^{(k+1)}(t,\mathbf{x}) = u_c^{(k)}(t,\mathbf{x}) + \varepsilon_{k+1} u_{k+1}(t,\mathbf{x}).
    \end{align}
\end{itemize}
\end{enumerate}

Unlike conventional regression problems, this approach targets the reduction of physics-based residuals rather than fitting directly function values (which are unknown for the problems we want to solve). This formulation enables progressive refinement of the solution, ultimately achieving better accuracy for challenging multi-scale problems such as the Taylor-Green vortex.

We apply this multi-stage procedure within each time window up to $T = 8$ for this problem. Specifically, we use a two-stage setup with prefactors $\epsilon_0 =1$ and $\epsilon_1 = 0.001$. 
However, we find it effective only when the first-stage network already achieves sufficiently low PDE residuals (e.g., $< 10^{-8}$). In cases where the initial PDE residual remains large, subsequent stages fail to effectively reduce the error. For   the Kolmogorov flow and turbulent channel flow, achieving this level of accuracy in the initial stage proves difficult. As a result, the multi-stage scheme offers limited improvement for these cases and is not employed.

\clearpage
\section{Computational Costs}
\label{appendix:cost}

All experiments are conducted using our JAX-based implementation, JAX-PI \cite{wang2023expert}, on a single NVIDIA H200 GPU. Detailed runtime benchmarks are reported in Table~\ref{tab: cost}.

We acknowledge that the current training times of PINNs remain significantly higher than those of conventional numerical solvers. This performance gap highlights a critical challenge and simultaneously motivates the development of next-generation physics-informed architectures and training strategies. Despite their computational overhead, our results demonstrate that PINNs can achieve stable and accurate simulations for complex turbulent flows in a data-free setting -- an important step toward practical and scalable scientific machine learning.

\begin{table}[h]
\centering
\renewcommand{\arraystretch}{1.4}
\caption{\textbf{Training time per time window (in hours)} for each benchmark problem using PINNs with the SOAP optimizer. All experiments are conducted on a single NVIDIA H200 GPU.}
\label{tab: cost}
\begin{tabular}{l c}
\toprule
\textbf{Benchmark} & \textbf{Training Time (per time window)} \\
\midrule
2D Kolmogorov flow & 3.00 \\
3D Taylor–Green vortex & 7.45 \\
3D Turbulent channel flow & 7.20 \\
\bottomrule
\end{tabular}
\vspace{1mm}
\end{table}

\section{Additional Visualizations}
\label{appendix:visualizations}

\subsection{2D Turbulent  Kolmogorov Flow}

\begin{figure}[htpb]
    \centering
    \includegraphics[width=1.0\linewidth]{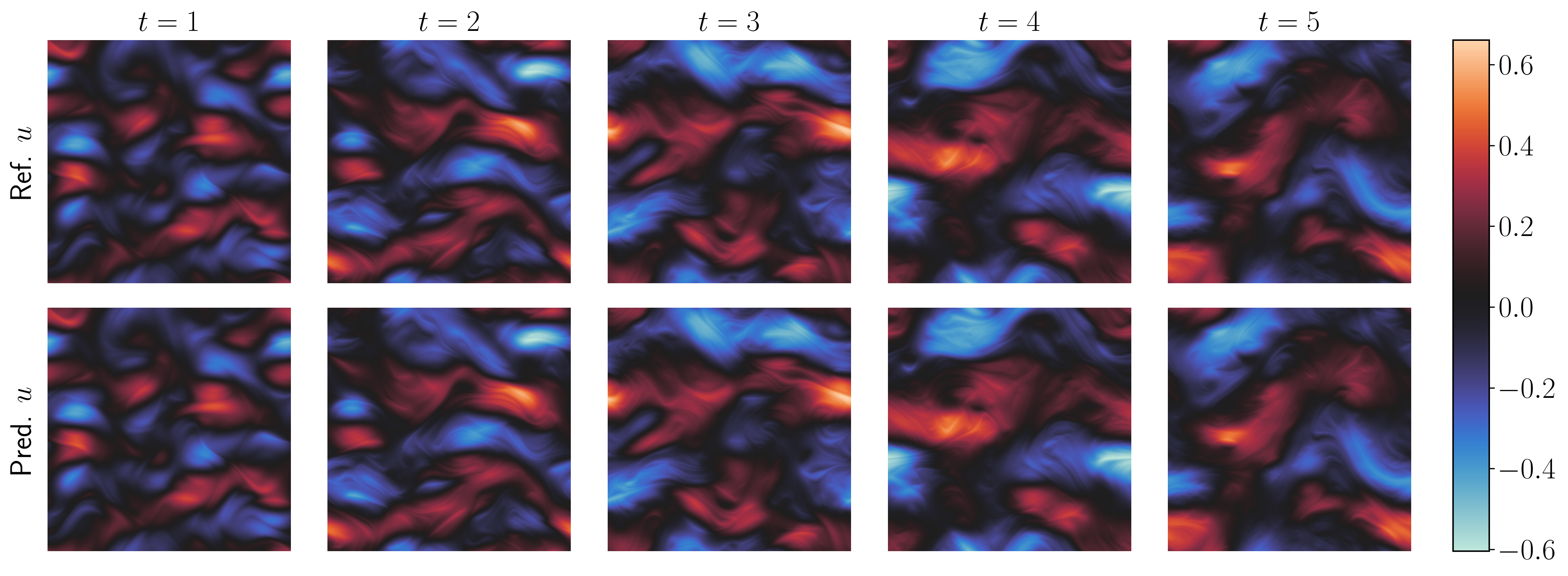}
    \caption{{\em 2D Turbulent  Kolmogorov flow.} Comparison of the predicted velocity $u$ against the corresponding numerical reference  at different time snapshots.}
    \label{fig:kf_u}
\end{figure}

\begin{figure}[htpb]
    \centering
    \includegraphics[width=1.0\linewidth]{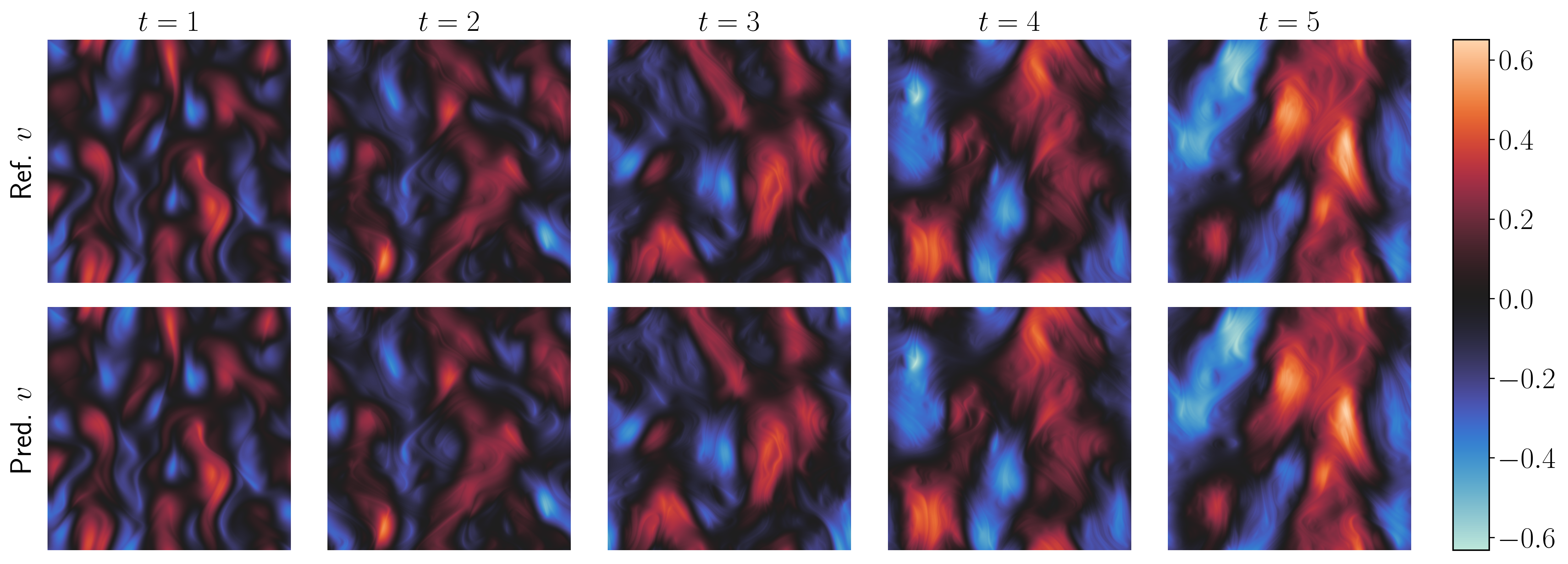}
    \caption{{\em 2D Turbulent  Kolmogorov flow.} Comparison of the predicted velocity $v$ against the corresponding numerical reference  at different time snapshots.}
    \label{fig:kf_v}
\end{figure}

\begin{figure}[htbp]
    \centering
    \includegraphics[width=1.0\linewidth]{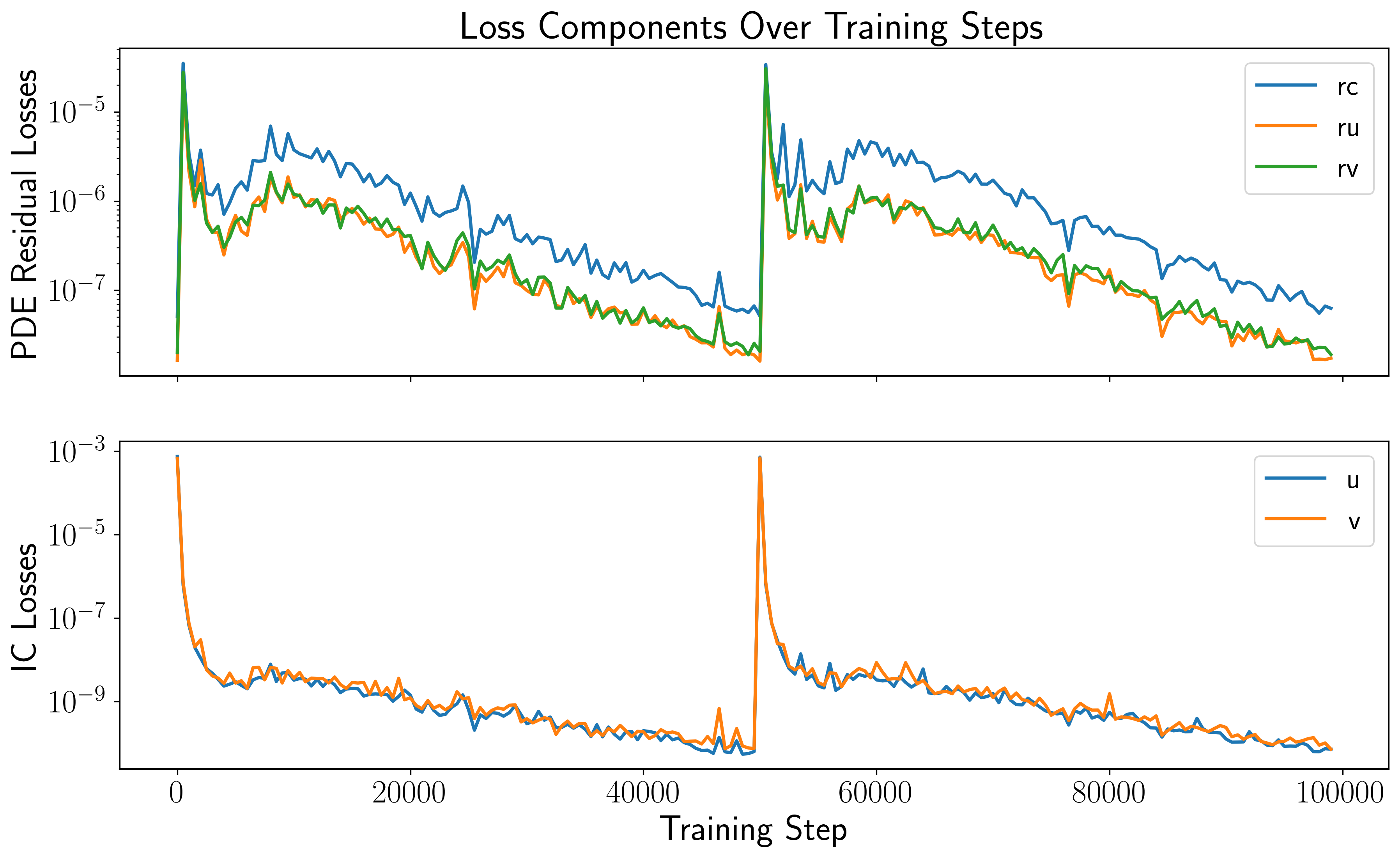}
  \caption{{\em 2D Turbulent  Kolmogorov flow.} Evolution of loss values across two consecutive time windows. The sharp spikes in loss values around step $5 \times 10^4$ mark the transition to training the next time window. IC loss denotes the initial condition loss,  including \texttt{u} and \texttt{v}  components.     The PDE residual losses include \texttt{ru}, \texttt{rv}, and \texttt{rc}, corresponding to the momentum equations in the $u$, $v$, directions and the continuity equation, respectively.}
    \label{fig:kf_loss}
\end{figure}

\clearpage
\subsection{3D Taylor-Green Vortex}

\begin{figure}[htbp]
    \centering
    \includegraphics[width=0.9\linewidth]{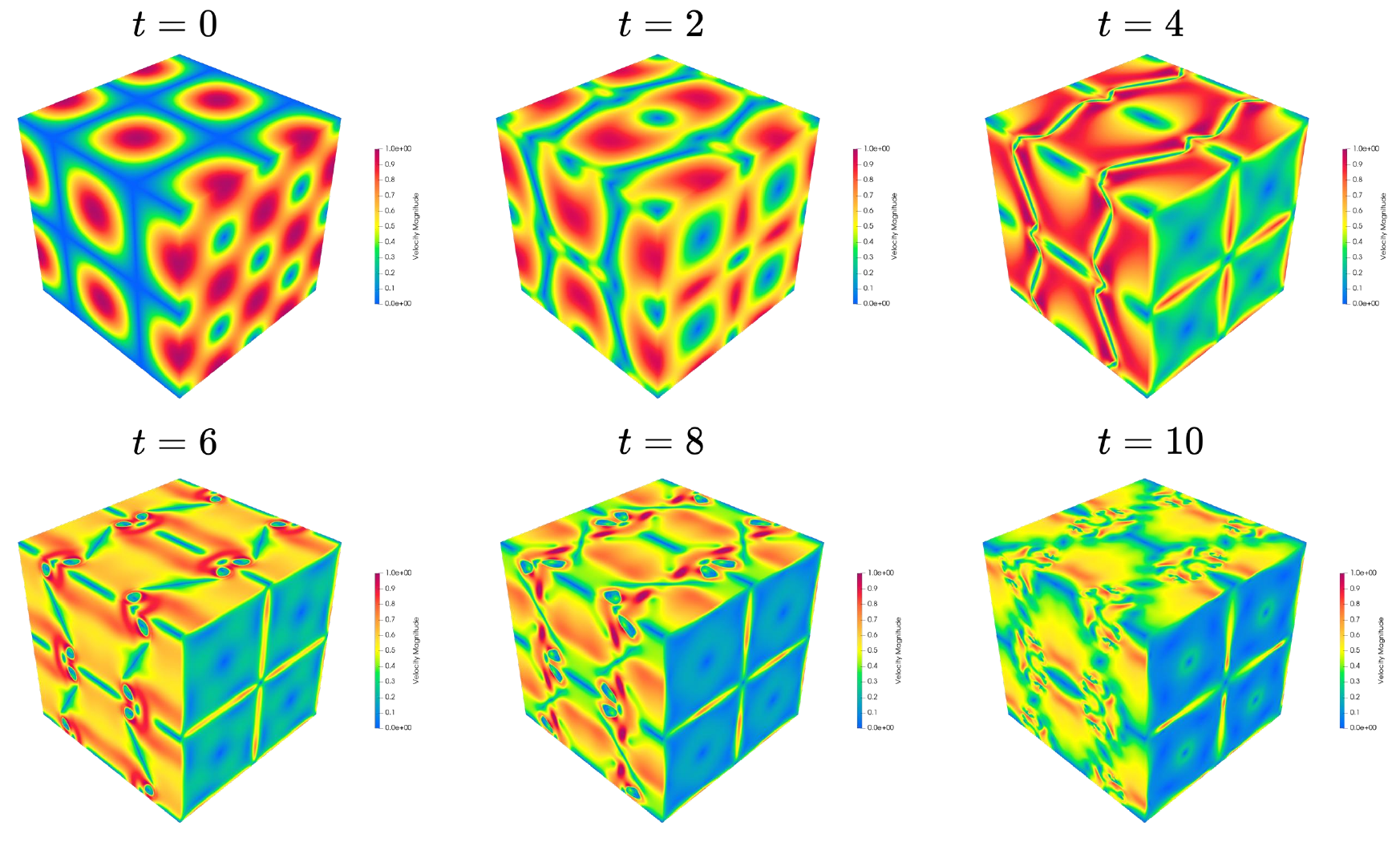}
    \caption{{\em Taylor-Green Vortex.} Predicted velocity norm at different time snapshots.}
    \label{fig:tgv_vel}
\end{figure}

\begin{figure}[htbp]
    \centering
    \includegraphics[width=0.9\linewidth]{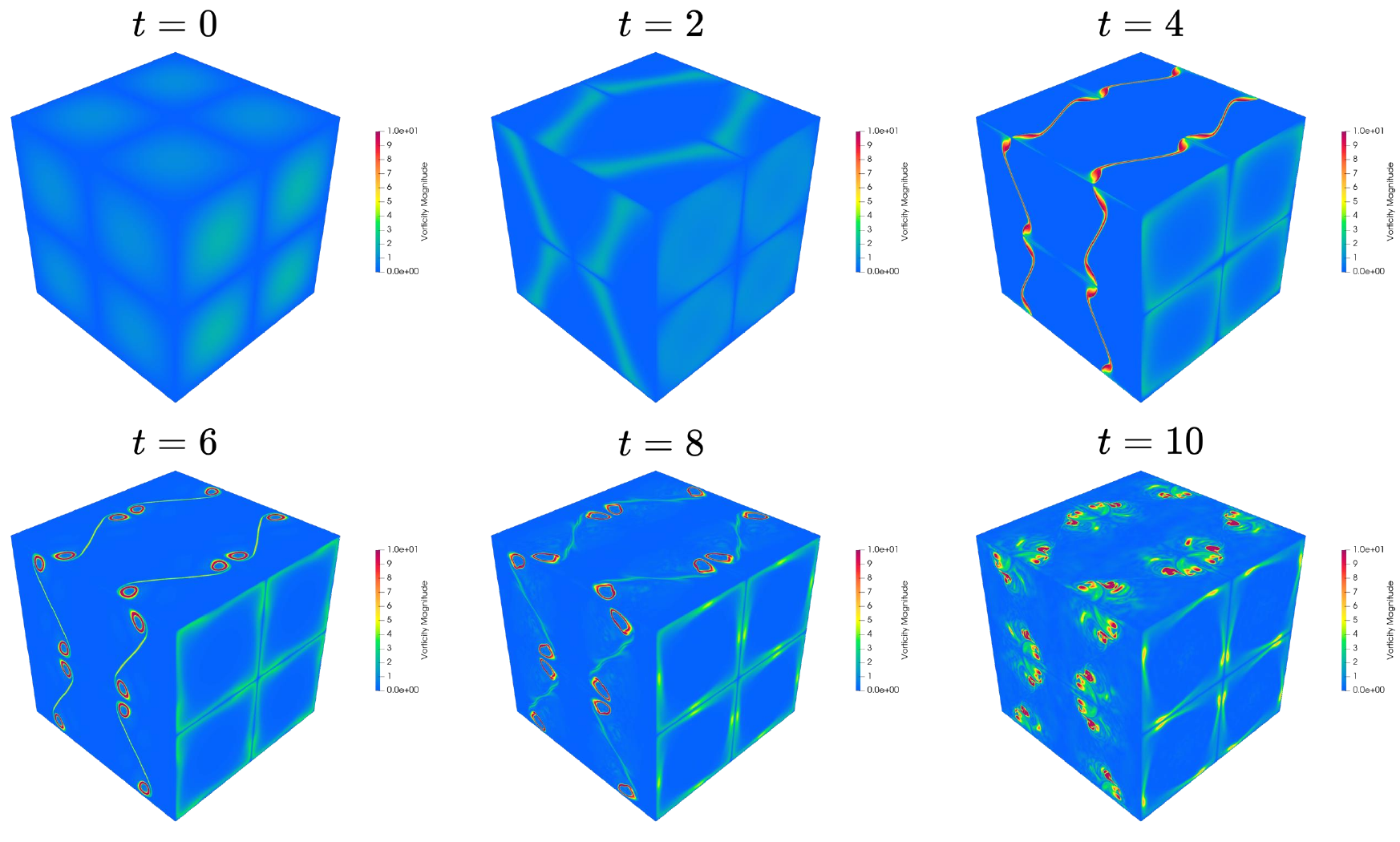}
    \caption{{\em Taylor-Green Vortex.} Predicted vorticity norm at different time snapshots.}
    \label{fig:tgv_vor}
\end{figure}

\begin{figure}[htbp]
    \centering
    \includegraphics[width=1.0\linewidth]{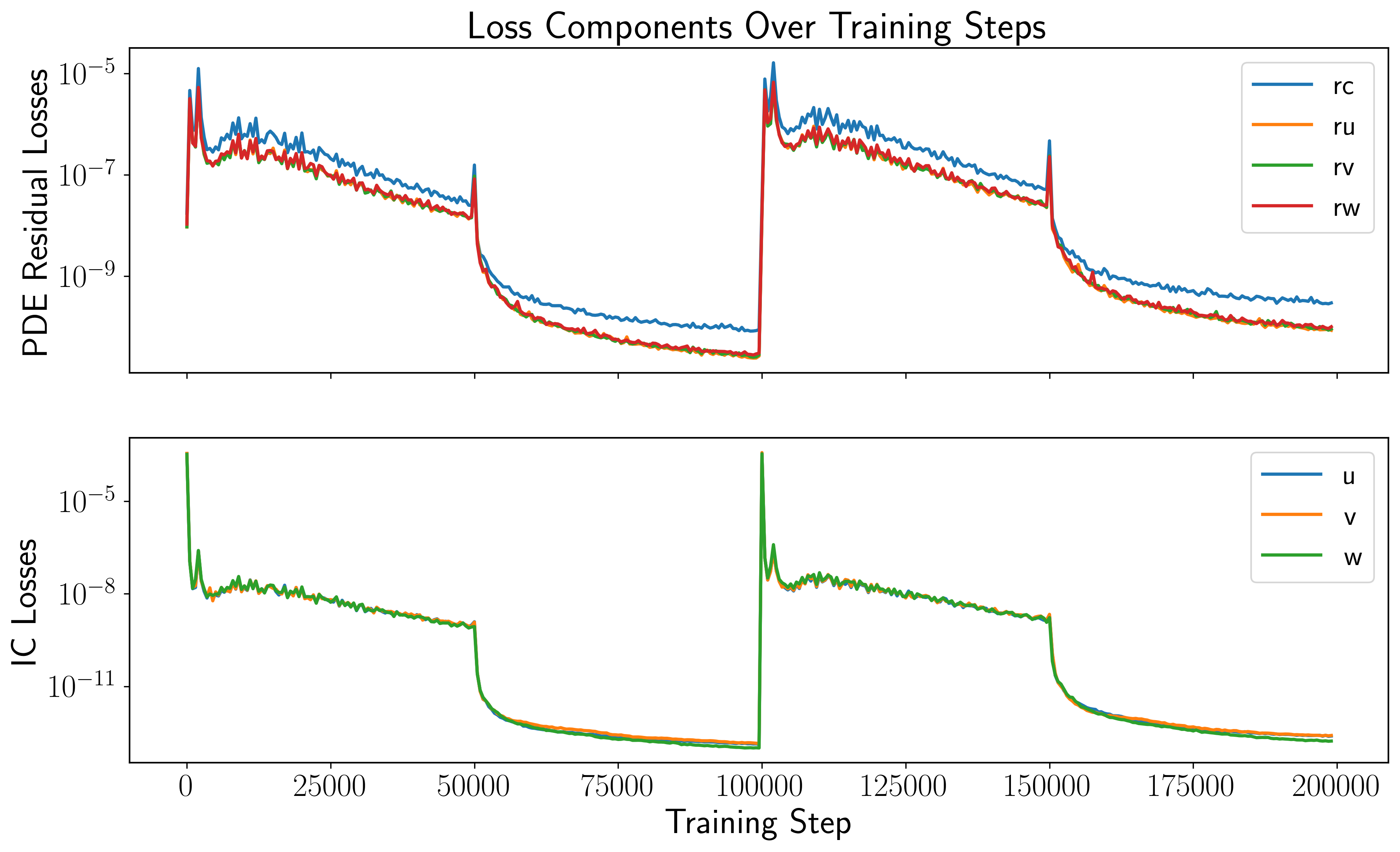}
   \caption{{\em Taylor-Green Vortex.} Evolution of loss values over two consecutive time windows. The sharp spikes in loss around step $\sim 10^5$ indicate the transition to the next time window, while the sudden drop at $5 \times 10^4$ corresponds to the start of the second training stage within the same window. IC loss denotes the initial condition loss,  including \texttt{u}, \texttt{v}, \texttt{w} components.   
   The PDE residual losses include \texttt{ru}, \texttt{rv}, \texttt{rw}, and \texttt{rc}, corresponding to the momentum equations in the $u$, $v$ and $w$ directions, and the continuity equation, respectively.}
    \label{fig:tgv_loss}
\end{figure}

\clearpage
\subsection{3D Turbulent Channel Flow}

\begin{figure}[htbp]
    \centering
    \includegraphics[width=1.0\linewidth]{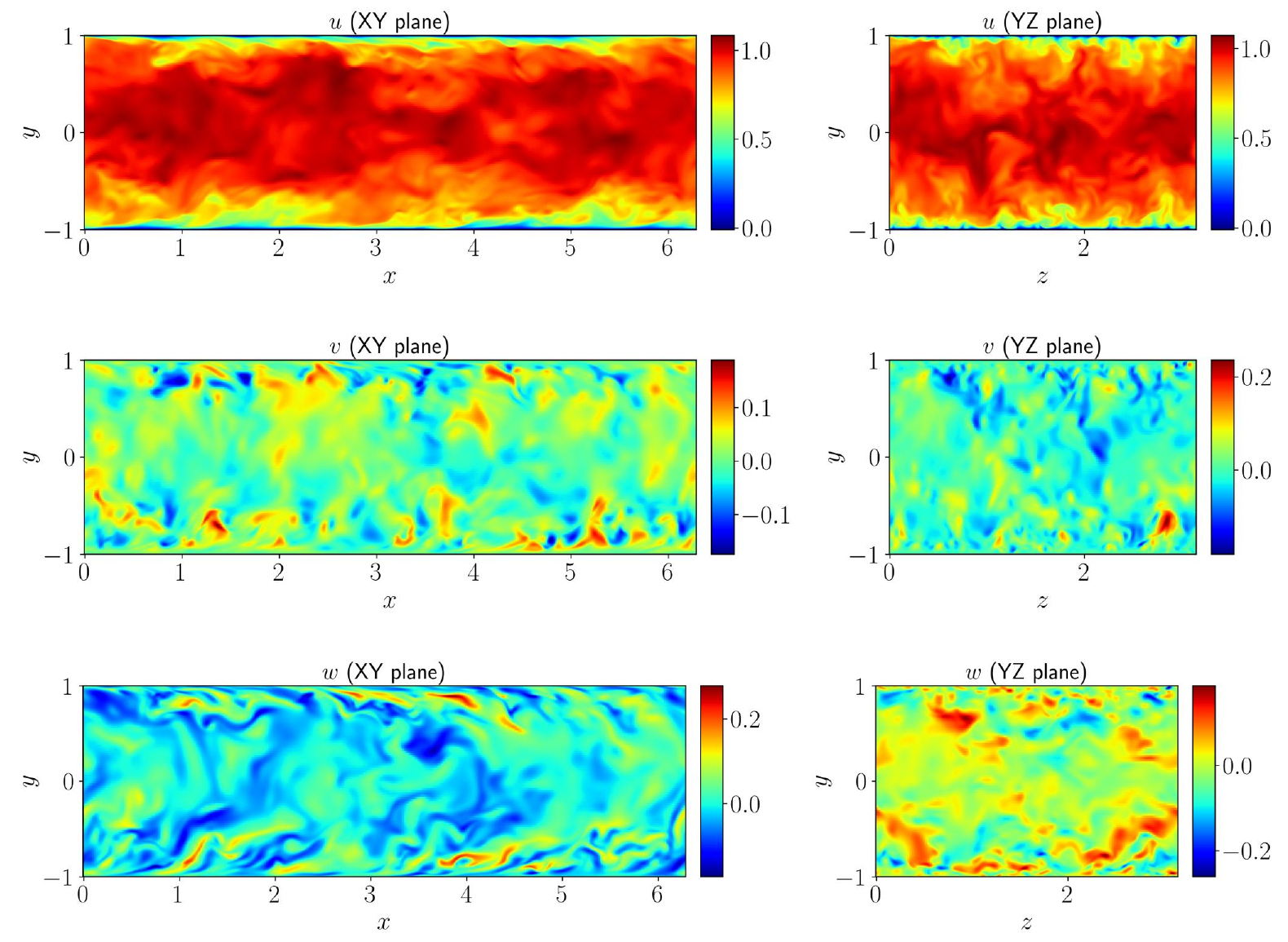}
   \caption{{\em Turbulent channel flow.} Predicted velocity field slices in the XY and YZ planes.}
    \label{fig:tcf_vel_slices}
\end{figure}

\begin{figure}[htbp]
    \centering
    \includegraphics[width=1.0\linewidth]{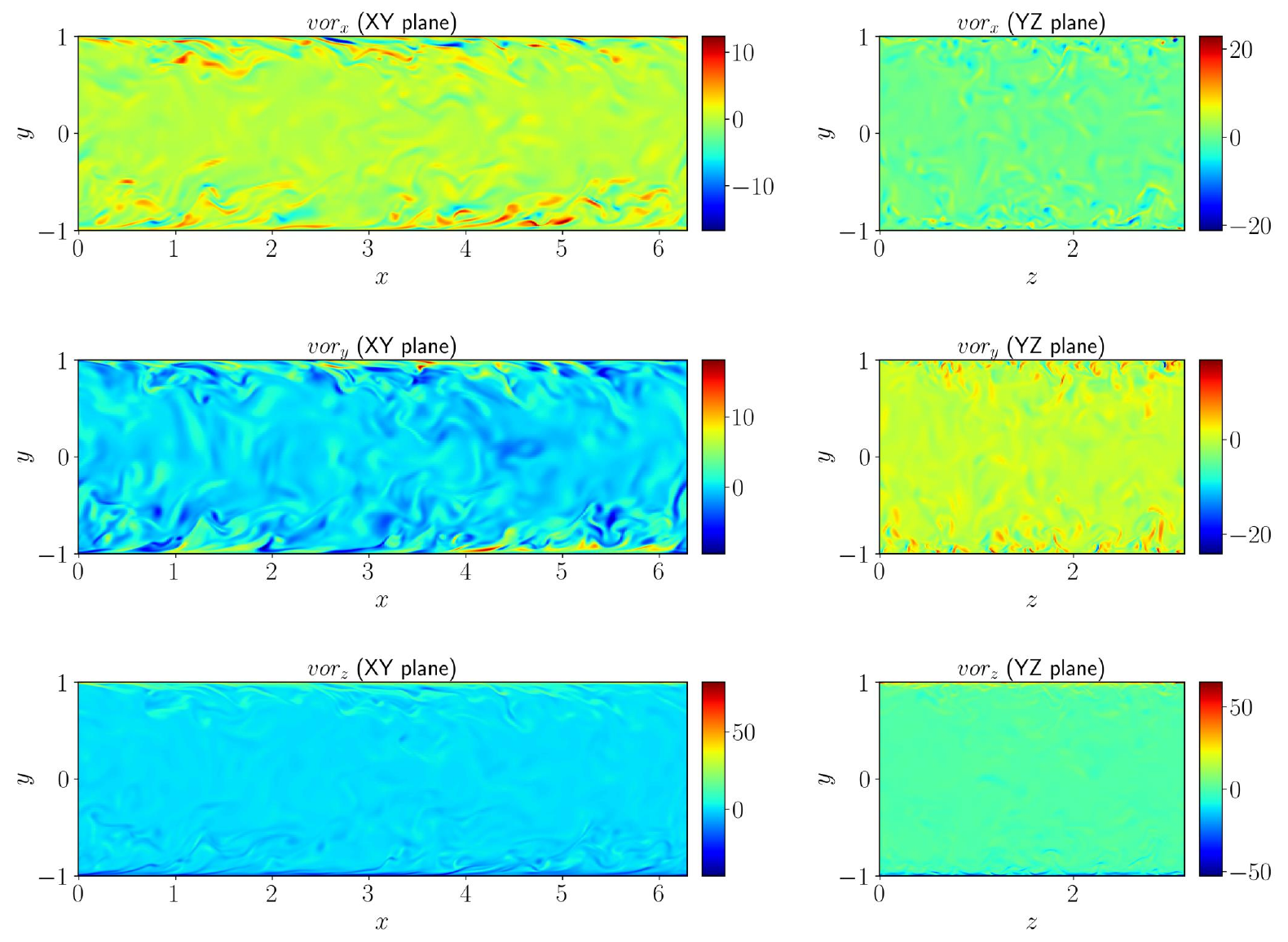}
    \caption{{\em Turbulent channel flow.} Predicted vorticity field slices at XY and YZ planes.}
    \label{fig:tcf_vor_slices}
\end{figure}

\begin{figure}[htbp]
    \centering
    \includegraphics[width=0.5\linewidth]{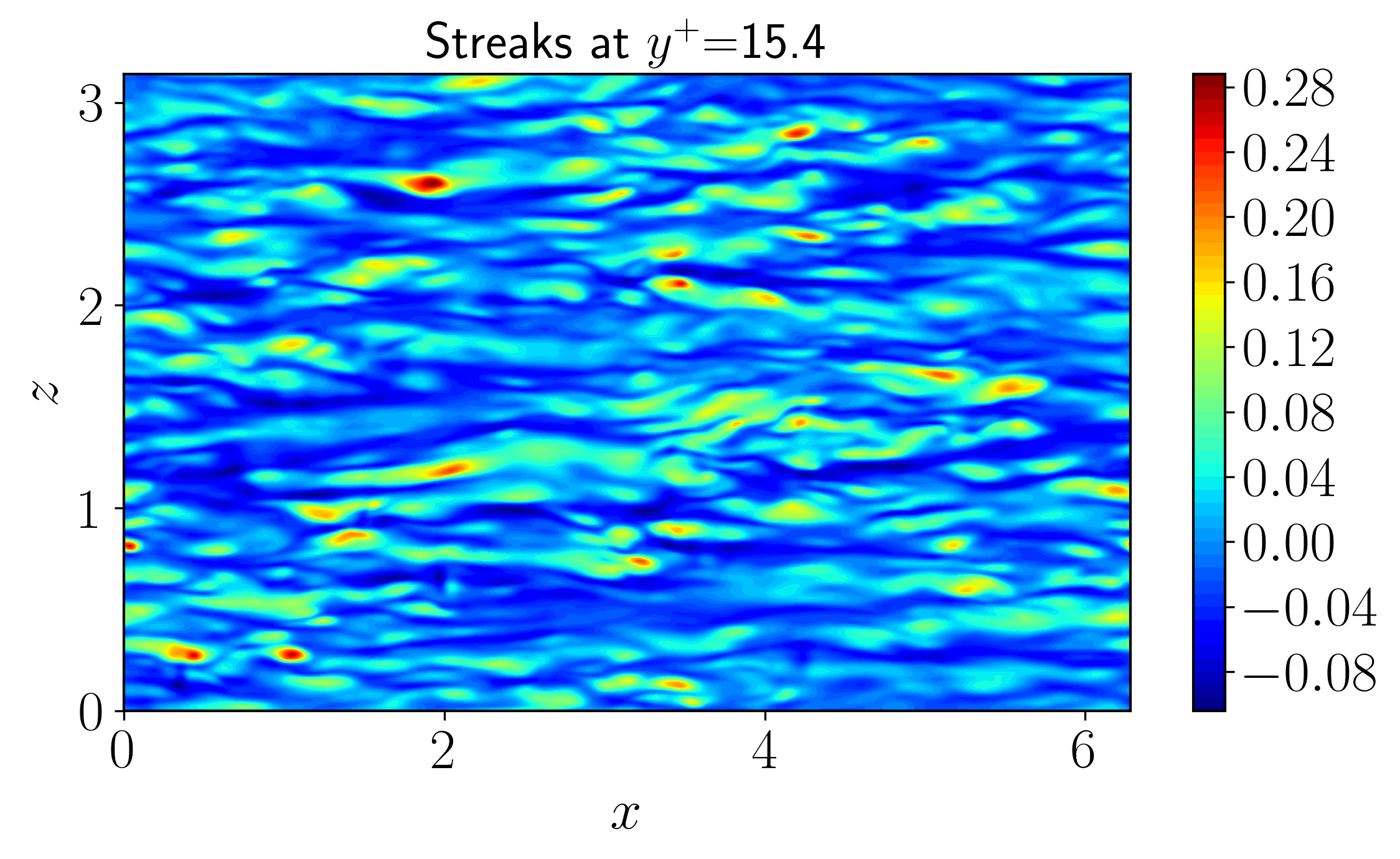}
    \caption{{\em Turbulent channel flow.} Predicted near-wall streaks of streamwise velocity fluctuations, visualized in the $x$–$z$ plane.}
    \label{fig:tcf_streaks}
\end{figure}

\begin{figure}[htbp]
    \centering
    \includegraphics[width=1.0\linewidth]{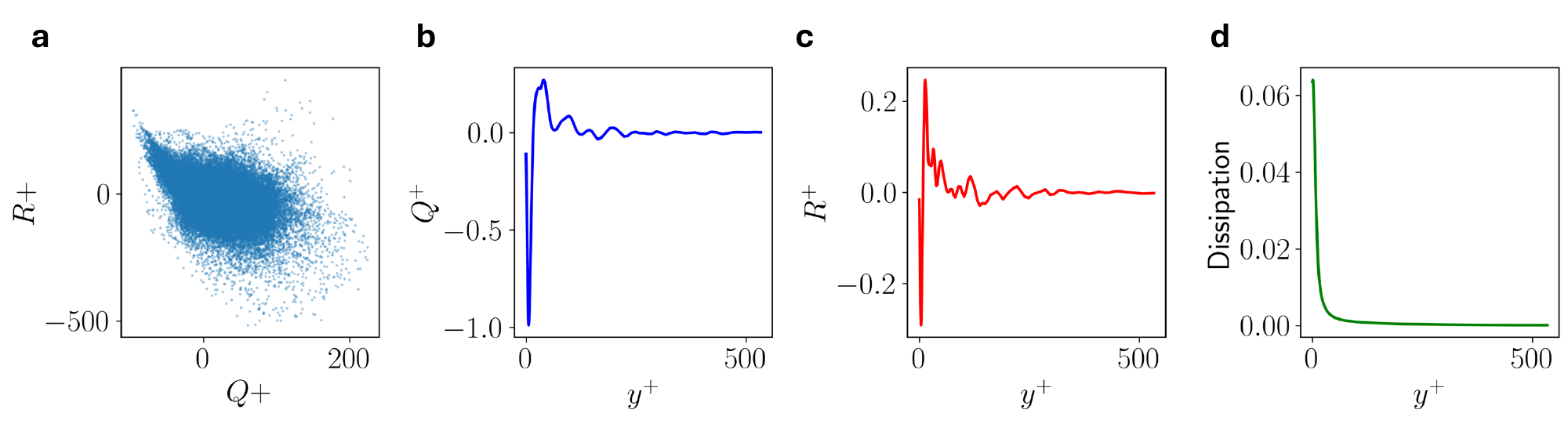}
    \caption{{\em Turbulent channel flow.} (a) Joint distribution of the invariants $Q^+$ and $R^+$. (b) Wall-normal profile of the mean invariant $Q^+$. (c) Wall-normal profile of the mean invariant $R^+$. (d) Wall-normal profile of the mean dissipation rate per unit mass $\epsilon^+$. All quantities are normalized in wall units.}
    \label{fig:tcf_other_stats}
\end{figure}

\begin{figure}
    \centering
    \includegraphics[width=1.0\linewidth]{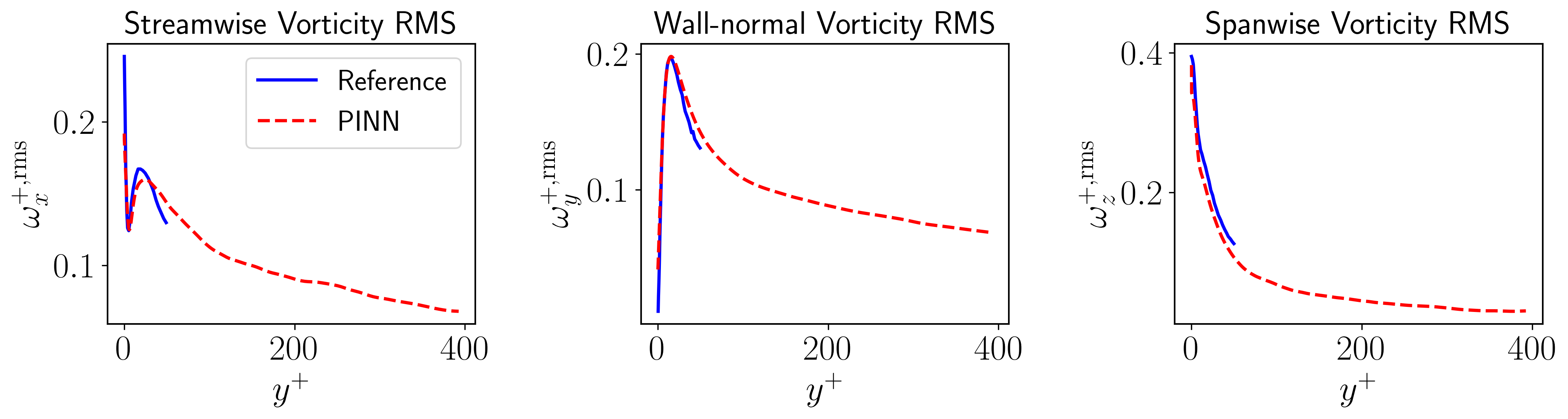}
    \caption{{\em Turbulent channel flow.} Mean profiles of the root-mean-square (rms) vorticity components $\omega_x^+$, $\omega_y^+$, and $\omega_z^+$ across the wall-normal direction, comparing reference DNS data and the PINN predictions. Note that the reference data from~\cite{moser1999direct} only reports the vorticity in the near wall region ($y^+ \in [0,50]$).}
    \label{fig:tcf_mean_vorticity}
\end{figure}

\begin{figure}
    \centering
    \includegraphics[width=1.0\linewidth]{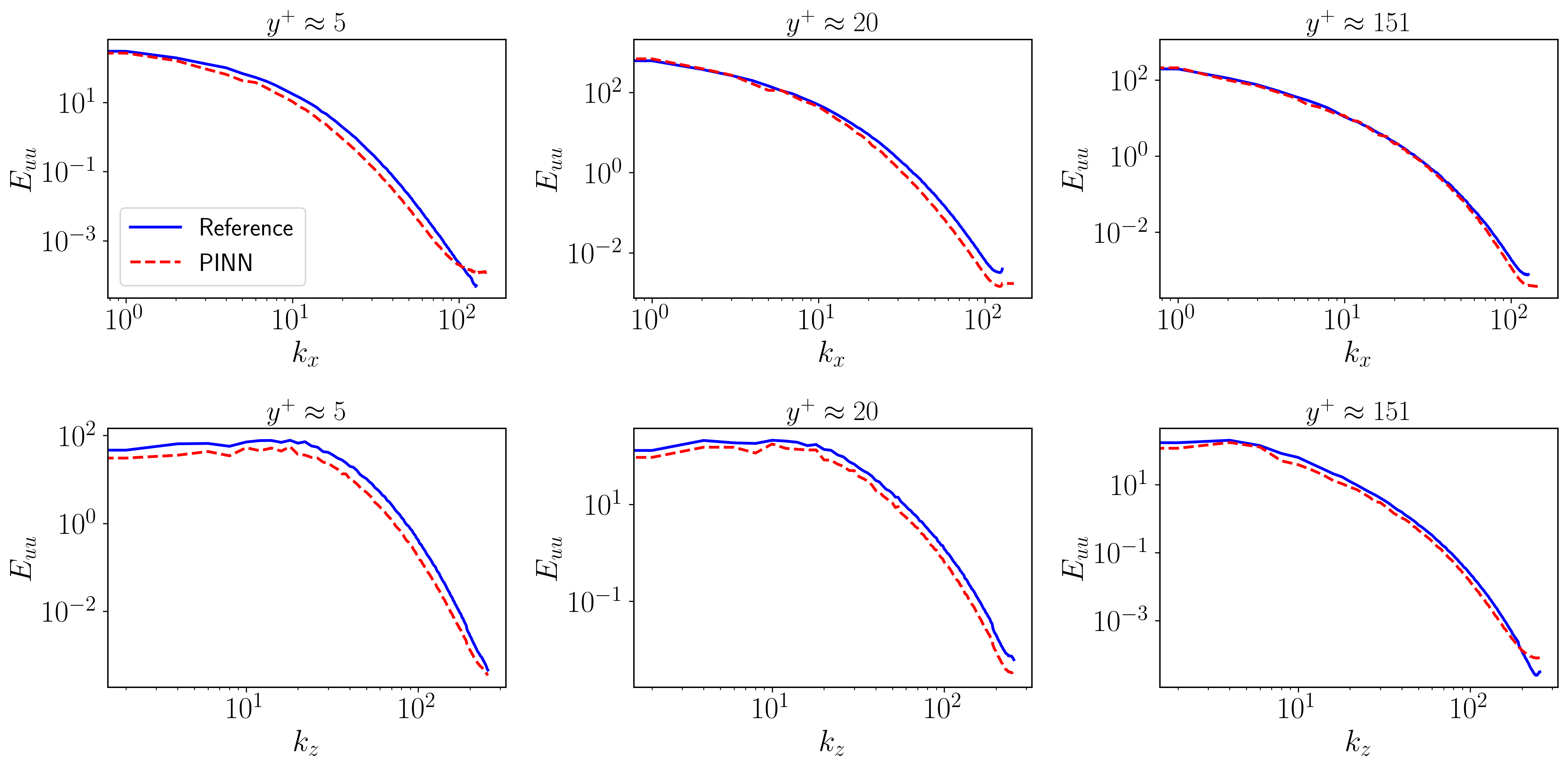}
\caption{{\em Turbulent channel flow.} Streamwise velocity energy spectrum $E_{uu}$ along streamwise direction ($k_x$) and spanwise direction ($k_z$), comparing reference DNS data with PINN predictions.}
    \label{fig:energy_spectra_uu}
\end{figure}

\begin{figure}
    \centering
    \includegraphics[width=1.0\linewidth]{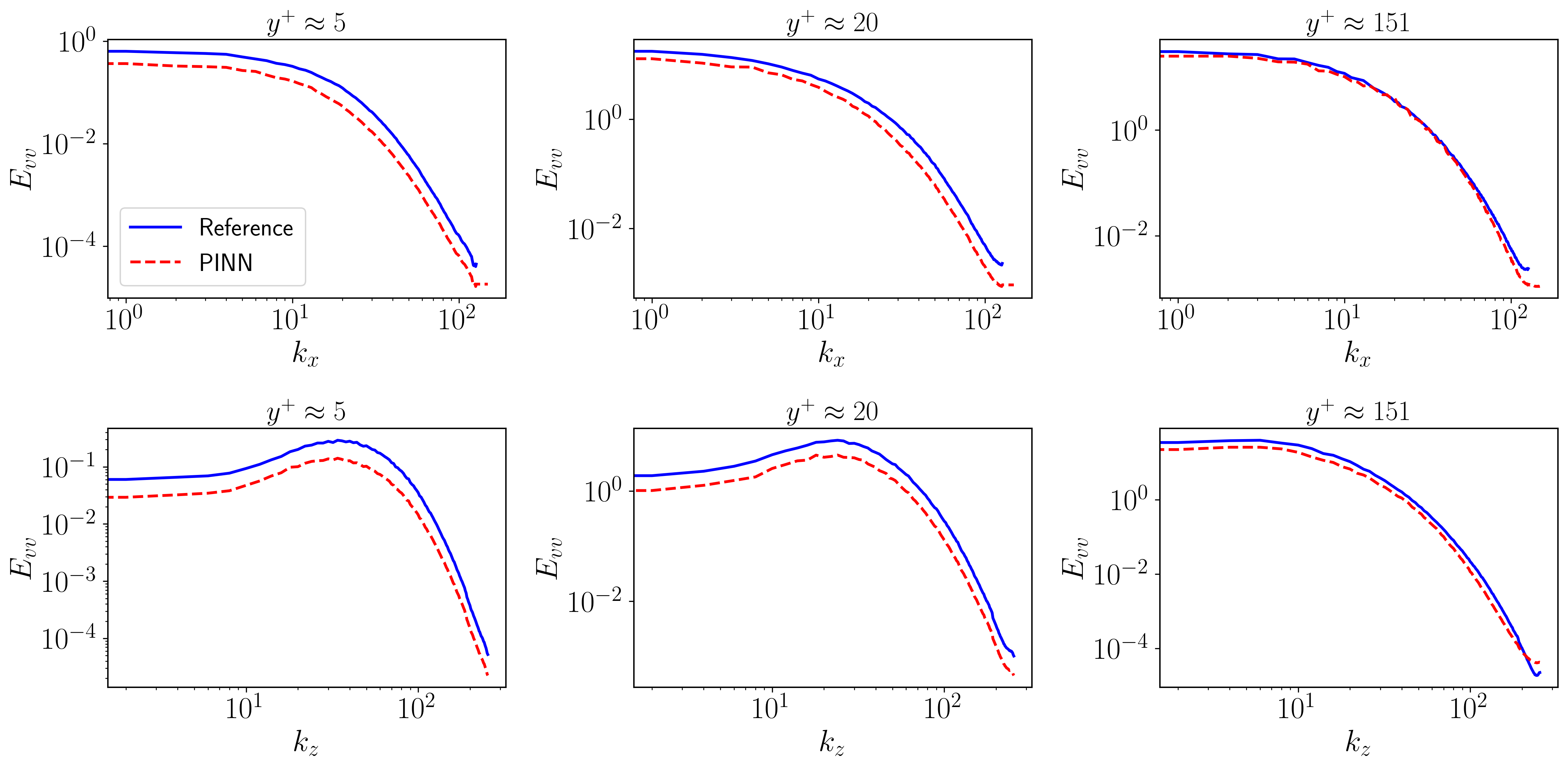}
\caption{{\em Turbulent channel flow.} Wall-normal velocity energy spectrum $E_{vv}$ along streamwise direction ($k_x$) and spanwise direction ($k_z$), comparing reference DNS data with PINN predictions.}
    \label{fig:energy_spectra_vv}
\end{figure}

\begin{figure}
    \centering
    \includegraphics[width=1.0\linewidth]{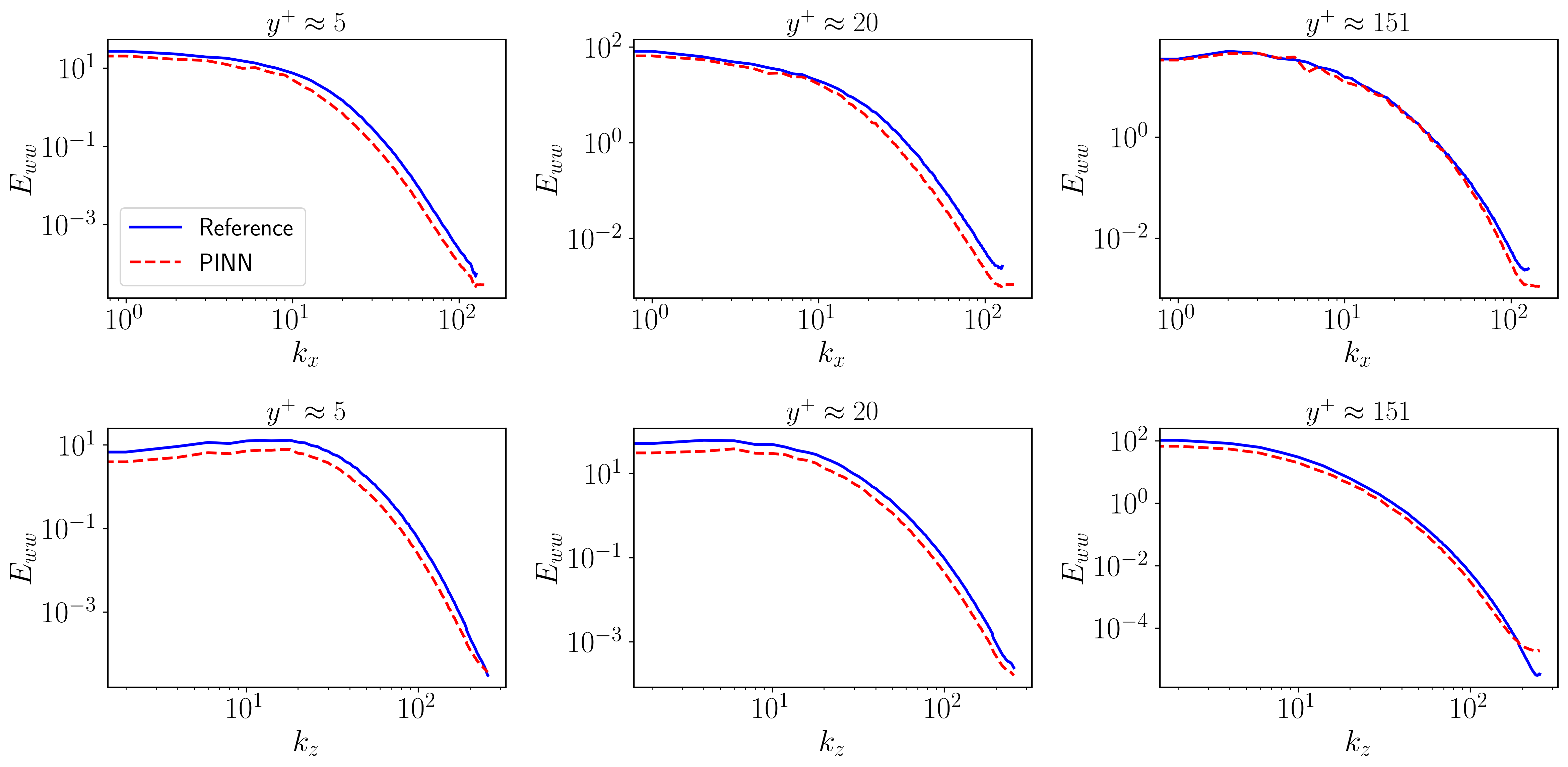}
\caption{{\em Turbulent channel flow.} Spanwise velocity energy spectrum $E_{ww}$ along streamwise direction ($k_x$) and spanwise direction ($k_z$), comparing reference DNS data with PINN predictions.}
    \label{fig:energy_spectra_ww}
\end{figure}

\begin{figure}[htbp]
    \centering
    \includegraphics[width=1.0\linewidth]{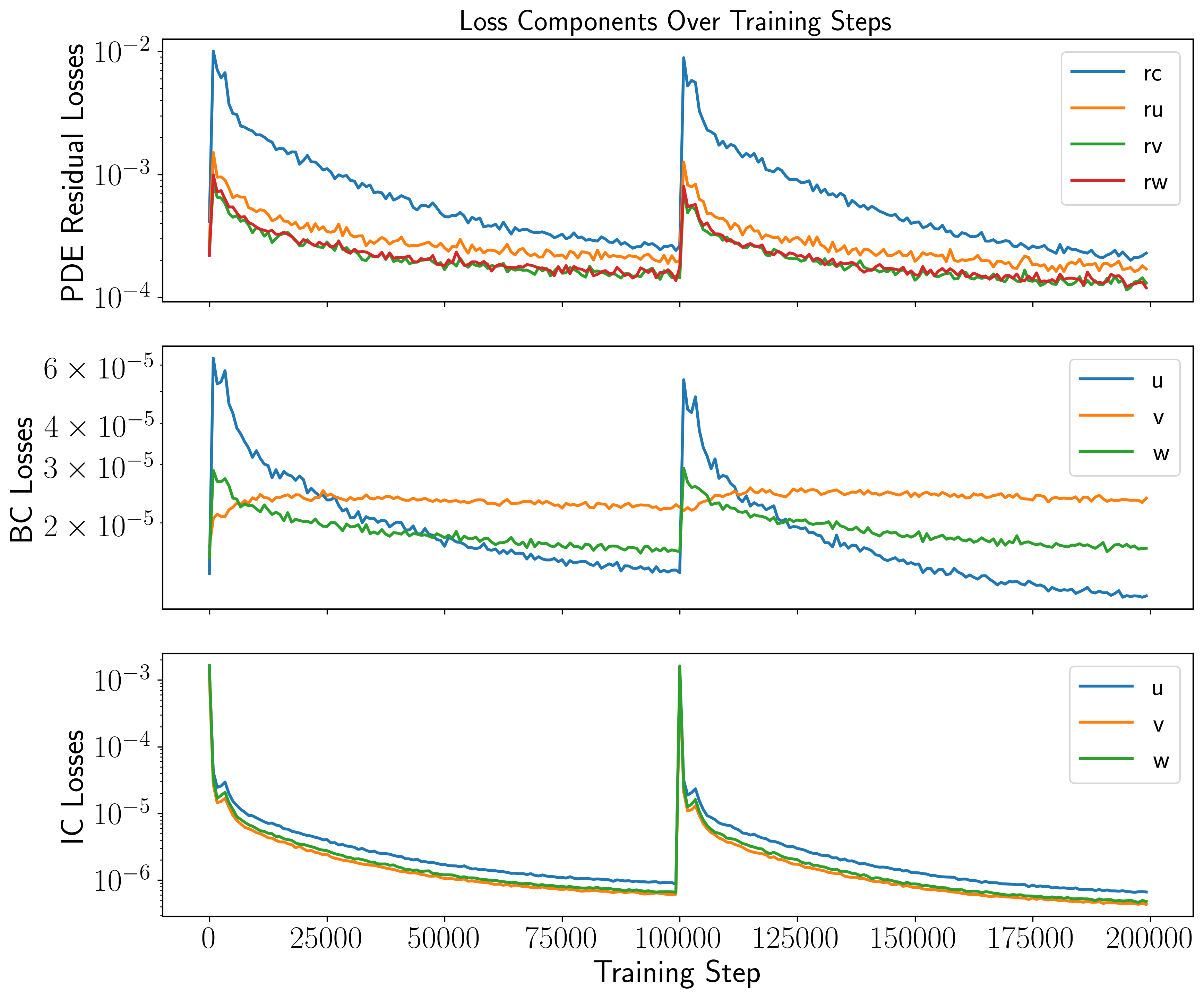}
    \caption{{\em Turbulent channel flow.} Evolution of loss values across two consecutive time windows. The sharp spikes in loss values around step $10^5$ mark the transition to training the next time window. IC loss refers to initial condition loss, and BC loss denotes boundary condition loss. Each include \texttt{u}, \texttt{v}, \texttt{w} components. 
    The PDE residual losses include \texttt{ru}, \texttt{rv}, \texttt{rw}, and \texttt{rc}, corresponding to the momentum equations in the $u$, $v$, $w$ directions and the continuity equation, respectively.}
    \label{fig:tcf_loss}
\end{figure}

\end{document}